\documentclass{article}

\usepackage{microtype}
\usepackage{graphicx}
\usepackage{subfigure}
\usepackage{booktabs} 

\usepackage{hyperref}

\usepackage[accepted]{icml2024}

\usepackage{amsmath}
\usepackage{amssymb}
\usepackage{mathtools}
\usepackage{amsthm}

\usepackage{amsmath,amsfonts,bm}

\def\eqref#1{equation~\ref{#1}}

\def\1{\bm{1}}

\def\mD{{\bm{D}}}

\def\mG{{\bm{G}}}

\DeclareMathAlphabet{\mathsfit}{\encodingdefault}{\sfdefault}{m}{sl}
\SetMathAlphabet{\mathsfit}{bold}{\encodingdefault}{\sfdefault}{bx}{n}

\newcommand{\E}{\mathbb{E}}

\makeatletter
\newcommand{\newreptheorem}[2]{%
\newenvironment{rep#1}[1]{%
 \def\rep@title{#2 \ref{##1}}%
 \begin{rep@theorem}}%
 {\end{rep@theorem}}}
\makeatother

\newreptheorem{theorem}{Theorem}

\newcommand{\printfnsymbol}[1]{%
  \textsuperscript{\@fnsymbol{#1}}%
}

\newcommand{\btheta}{\boldsymbol{\theta}}

\newcommand{\bgamma}{\boldsymbol{\gamma}}

\newcommand{\bphi}{\boldsymbol{\phi}}

\usepackage[capitalize,noabbrev]{cleveref}

\theoremstyle{plain}

\theoremstyle{definition}

\theoremstyle{remark}

\usepackage[textsize=tiny]{todonotes}

\icmltitlerunning{Challenges and Considerations in the Evaluation of Bayesian Causal Discovery}

\begin{document}

\twocolumn[
\icmltitle{Challenges and Considerations in the Evaluation of Bayesian Causal Discovery}



\icmlsetsymbol{equal}{*}

\begin{icmlauthorlist}
\icmlauthor{Amir Mohammad Karimi Mamaghan}{kth}
\icmlauthor{Panagiotis Tigas}{oatml}
\icmlauthor{Karl Henrik Johansson}{kth,dig}
\icmlauthor{Yarin Gal}{oatml}
\icmlauthor{Yashas Annadani}{equal,helmholtz,tum}
\icmlauthor{Stefan Bauer}{equal,helmholtz,tum}
\end{icmlauthorlist}

\icmlaffiliation{kth}{KTH Royal Institute of Technology}
\icmlaffiliation{oatml}{OATML, University of Oxford}
\icmlaffiliation{dig}{Digital Futures}
\icmlaffiliation{helmholtz}{Helmholtz AI}
\icmlaffiliation{tum}{TU Munich}

\icmlcorrespondingauthor{Amir Mohammad Karimi Mamaghan}{amkm@kth.se}

\icmlkeywords{Bayesian Causal Discovery, Approximate Inference}

\vskip 0.3in
]



\printAffiliationsAndNotice{\icmlEqualContribution} 
\begin{abstract}

\looseness=-1 Representing uncertainty in causal discovery is a crucial component for experimental design, and more broadly, for safe and
reliable causal decision making. Bayesian Causal Discovery (BCD) offers a principled approach to encapsulating this uncertainty. Unlike non-Bayesian causal discovery, which relies on a single estimated causal graph and model parameters for assessment, evaluating BCD presents challenges due to the nature of its inferred quantity – the posterior distribution. As a result, the research community has proposed various metrics to assess the quality of the approximate posterior. However, there is, to date, no consensus on the most suitable metric(s) for evaluation. In this work, we reexamine this question by dissecting various metrics and understanding their limitations. Through extensive empirical evaluation, we find that many existing metrics fail to exhibit a strong correlation with the quality of approximation to the true posterior, especially in scenarios with low sample sizes where BCD is most desirable. We highlight the suitability (or lack thereof) of these metrics under two distinct factors: the identifiability of the underlying causal model and the quantity of available data. Both factors affect the entropy of the true posterior, indicating that the current metrics are less fitting in settings of higher entropy. Our findings underline the importance of a more nuanced evaluation of new methods by taking into account the nature of the true posterior, as well as guide and motivate the development of new evaluation procedures for this challenge.
\end{abstract}

\section{Introduction}
\label{sec:intro}

Much of the pursuit in acquiring scientific knowledge involves inferring causal relationships within a system of interest and the laws that govern those relationships. Applications in biology, like inferring a gene network and their regulatory mechanisms from gene expression data~\citep{tejada2023causal, Dibaeinia2020SERGIOAS} and protein-signaling networks with single-cell data~\citep{sachs2005causal},  necessitates a mechanistic understanding of the data generation process. Estimating the causal model in such applications from data, called causal discovery, is an important problem in empirical sciences~\citep{spirtes2000causation}. 

A typical scientific discovery loop for causal understanding involves a scientist first coming up with causal hypotheses based on prior knowledge, and refining these hypotheses based on new evidence obtained through observation and experimentation. In science, a key requirement in light of limited data is that all the plausible hypotheses that explain the data have to be considered before devising an efficient experimentation protocol, as opposed to a single most likely one~\citep{lindley1956measure,chaloner1995bayesian}. Bayesian Causal Discovery (BCD) is an elegant framework that fulfills this requirement by quantifying the epistemic uncertainty of the underlying causal model through the Bayesian posterior, which provides a degree of belief of every causal hypothesis proportional to its ability to explain the data~\citep{heckerman2006bayesian, friedman2003being, chickering2013bayesian}. The quantified epistemic uncertainty can be then used to design informative experiments/ interventions to perform~\citep{tong2001active,tigas2022interventions, sussex2022model, lyle2023discobax} or to estimate causal effects of variables with Bayesian model averaging~\citep{geffner2022deep,emezue2023benchmarking}.

One of the common frameworks for dealing with questions related to cause and effect is the Structural Causal Model (SCM) with an associated graph indicating the causal relationships between variables~\citep{pearl2009causality,peters2017elements}. Under this framework, BCD aims to infer the Bayesian posterior over the graph and the parameters of the SCM. This is a hard problem due to the combinatorial nature of graphs, which renders the posterior intractable for more than 6 variables. Recently, various approximate inference methods have been introduced that use gradient information allowing to scale to SCMs for which the true posterior is intractable~\citep{annadani2021variational,cundy2021bcd,lorch2021dibs,nishikawa2022bayesian,deleu2022bayesian,deleu2023joint,hagele2023bacadi,atanackovic2023dyngfn}. However, in the absence of the true posterior, evaluation of BCD methods is hard as the inferred quantity is a distribution rather than the most-likely estimate, as in causal discovery. The BCD community so far has relied on proxy metrics\footnote{Unless otherwise specified, metric(s) in this work refers to evaluation method(s) or protocol(s) to assess the goodness of an algorithm.}, many of which are derived from causal discovery evaluation. For instance, a standard metric in causal discovery is the Structural Hamming Distance (SHD) for evaluating the estimated graph, and in BCD, \emph{expected} SHD is used to evaluate the posterior over DAGs. However, such a metric may not be representative of how good the posterior approximation is, and has been briefly discussed in prior work~\citep{deleu2023joint,lorch2022amortized}. Our motivation is that a holistic understanding of the limitations of BCD evaluation is missing. Given that many new BCD algorithms are being proposed, proper understanding of the limitations of the present evaluation metrics is important to make advances in the right direction, especially with regards to applying it to real-world datasets where the amount of samples might be limited.

In this work, we aim to bridge the gap that exists in the understanding of the evaluation of BCD algorithms.  In order to do so, we note that the desiderata for an ideal evaluation metric would be to compare the approximated posterior to the \emph{gold standard}, the true posterior. Therefore, we analyze the performance of different BCD methods on all the known metrics for linear additive noise models, for which the true posterior is tractable. This not only helps us to compare how different metrics correlate with evaluating the approximate posterior against the true posterior, but also gives a way to understand some properties of the true posterior, which we shall show, is important in understanding the conditions under which the current metrics are suitable for evaluation or where they may be lacking. Our experimental evaluation with linear additive noise models on 8 different metrics for 5 different algorithms reveals the following aspects:
\begin{enumerate}
    \item In terms of the relative performance of BCD methods, we find that all the metrics are not correlated to a metric which directly evaluates on the true posterior when the number of samples is low ($n\approx d$ where $n$ is the dataset size and $d$ is the number of variables), indicating that the current metrics are not suitable for evaluation of uncertainty in these settings.
    \item With higher number of samples ($n >> d$), the correlation between the current evaluation metrics and the metric on the true posterior significantly improves.
    \item Based on a similar correlation analysis, we observe that the current metrics are less suitable for the evaluation of uncertainty when the true model is non-identifiable, as opposed to the identifiable case.
    \item Overall, the reliability of existing metrics as evaluation methods is related to the entropy of the true posterior. The true posterior has higher entropy with less data and non-identifiability. 
    \item Therefore, future algorithms should consider the setting of interest (and the entropy of the true posterior it induces) in deciding whether to use existing metrics or not. In a higher entropy true posterior setting, it would be better to evaluate the posterior on downstream tasks (for example causal effect estimation) where the ground truth is well-defined.
\end{enumerate}

The remaining parts of the paper are organized as follows: \cref{sec:background} provides background on causality and Bayesian causal discovery. \cref{sec:metrics} explains all different evaluation metrics for BCD in use and discusses their limitations. \cref{sec:exp} presents the empirical evaluation of multiple different algorithms on all metrics which highlights the shortcomings of present metrics for BCD evaluation in terms of the quality of the posterior approximation. \cref{sec:alternative_evaluation} proposes two alternative ways of evaluating BCD models. Finally, \cref{sec:conclusion} discusses the limitations and presents the overall conclusion.

\section{Background}
\label{sec:background}
In this section, we briefly introduce the Structural Causal Model (SCM) formalism under which the problem of causal discovery is defined. We also introduce the problem of Bayesian Causal Discovery under this framework.
\subsection{Structural Causal Model}
Let $\mathbf{V}=\{1,\dots,d\}$ be the vertex set of any graph $\mG=(\mathbf{V},E)$ and $\mathbf{X} = \{ \mathrm{X}_1, \dots, \mathrm{X}_d \} \subseteq \mathcal{X}$ be the random variables of interest indexed by $\mathbf{V}$. A Structural Causal Model (SCM) consists of a set of equations wherein each variable $X_i$ is assigned a value which is a deterministic function of its direct causes $X_{\text{pa}(i)}$ as well as an exogenous noise variable $\epsilon_i$ with a distribution $P_{\epsilon_i}$:
\begin{equation}
\label{eq:scm}
X_i \coloneqq f_i(X_{\text{pa}(i)}, \epsilon_i) \,\,\,\, \forall i\in \mathbf{V}  
\end{equation}
$f_i$'s are mechanisms that relate how the direct causes affect the variable $X_i$. If the structural assignments are assumed to be acyclic, these equations induce a Directed Acyclic Graph (DAG) $\mG=(\mathbf{V},E)$ whose vertices correspond to the variables and edges indicate direct causes. A perfect intervention on any variable $X_i$ corresponds to changing the structural equation of that variable to the desired state (value), $X_i\coloneqq s_i$, where $s_i\in \mathcal{X}_i$. It is denoted by the $\mathrm{do}$-operator~\citep{pearl2009causality} as $\mathrm{do}(X_i=s_i)$. Under this model, the recursive application of \cref{eq:scm} entails a joint distribution $p_\mathbf{X}$,
such that the Markov factorization holds:
\begin{equation}
  p_\mathbf{X}(\mathbf{X}) = \prod_{i=1}^d p_i(X_i|X_{\text{pa}(i)})  
\end{equation}
The problem of causal discovery is to estimate the SCM (i.e. the causal graph $\mG$, parameters of $f_i$'s and $\epsilon_i$'s) given $N$ samples from $p_\mathbf{X}$. For analysis of different evaluation metrics, we assume that the SCM is causally sufficient, i.e. all the variables are measurable, and the noise variables are mutually independent.

Without further assumptions on the mechanisms and the noise, an SCM is not \emph{identifiable} from observational data, i.e. there could be multiple factorizations that can be consistent with a given joint distribution $p_\mathbf{X}$. One of the simplest identifiable setting is a linear Gaussian Additive Noise Model (ANM) with homoscedastic noise~\citep{peters2014identifiability}:
\begin{equation}\label{eq:scm_additive}
X_i \coloneqq  \bgamma_i^T X_{\text{pa}(i)} + \epsilon_i,\,\,\,\epsilon_i\sim \mathcal{N}(0, \sigma^2) \nonumber
\end{equation}
where the mechanisms $f_i$ are linear with parameter $\bgamma_i \in \mathbb{R}^{|\text{pa}(i)|}$. For notational brevity, henceforth we denote $\bphi = (\bgamma_1, \dots,\bgamma_d,\sigma^2)$ and all the parameters of interest with $\btheta=(\mG, \bphi)$. If the noise is heteroscedastic in the above model, under the assumption of faithfulness, it is only identifiable up to an equivalence class over graphs, called Markov Equivalence Class (MEC)~\citep{andersson1997characterization}. 
\subsection{Bayesian Causal Discovery}
  Given a dataset $\mD=\{\mathbf{X}^{(1)},\ldots,\mathbf{X}^{(N)}\}$, DAG $\mG$ and parameters $\bphi$, they induce a unique joint distribution $p(\mD,\bphi, \mG)$ with the prior $p(\mG,\bphi)$ and likelihood $p(\mD\vert \mG,\bphi)$ \citep{friedman2003being}. Bayesian causal discovery aims to infer the posterior\footnote{We refer this as true posterior to emphasize the difference with approximate posterior.} $p(\mG,\bphi\vert \mD)\propto p(\mD\vert \mG,\bphi)p(\mG,\bphi)$. A Bayesian method for causal discovery is preferable to model epistemic uncertainty about the model due to finite data. In addition, with choice of appropriate parameter priors~\citep{geiger2002parameter},  equivalence classes like MEC can be characterized in the case of non-identifiability. A crucial benefit of posterior inference in causal models is that it is helpful for downstream tasks like experimental design and cause-effect estimation with Bayesian model averaging. 
  However, the true posterior is not tractable for more than 6 variables. The true posterior is given by $p(\mG, \bphi|\mD)=\frac{p(\mD|\bphi,\mG)p(\mG,\bphi)} {\sum_\mG\int_\phi p(\mD, \mG, \bphi)}$. To calculate the true posterior, we need to calculate the summation over G which is infeasible as the number of possible DAGs grows super-exponentially w.r.t. number of variables ($\mathcal{O}(2^{d^2})$).
  The goal of BCD therefore is to find an approximate posterior $q(\mG, \bphi\vert \mD)$ that is close to the true posterior. 
  
\section{On Evaluation of BCD}
\label{sec:metrics}
Evaluating the goodness of posterior approximation $q(\mG, \bphi\vert \mD)$ in the absence of true posterior $p(\mG,\bphi\vert \mD)$ requires proxy metrics or downstream task evaluation. The BCD community so far has focused on proxy metrics which are mostly derived from causal discovery evaluation. The current metrics can be classified into two categories: graph-only evaluation metrics and full posterior evaluation metrics. 
\paragraph{Graph only evaluation metrics.} These metrics aim to evaluate the uncertainty quantified about graphs through the approximate posterior $q(\mG\vert \mD)$.
\begin{itemize}
    \item \textbf{$\E$-SHD}: Structural Hamming Distance (SHD) is a measure of number of edges that are to be added, removed, or reversed to get the ground truth graph from the estimated graph. Since we have a posterior distribution $q(\mG\mid \mD)$ over graphs, the \emph{expected} SHD is measured: 
    \begin{equation*}
    \E\text{-SHD} \coloneqq \E_{\mG\sim q(\mG\vert \mD)}[\mathrm{SHD}(\mG, \mG^{{GT}})]
\end{equation*} where $\mG^{GT}$ is the ground-truth causal graph. 
\item \textbf{$\E$-CPDAG SHD}: Similar to $\E$-SHD, $\E$-CPDAG SHD measures the expected hamming distance between the Completed partially directed acyclic graph (CPDAG, an equivalence class of DAGs) of the ground truth graph and the CPDAG of the graph sampled from the posterior. 
    
    \item Threshold Metrics: Area Under Precision Recall Curve (\textbf{AUPRC}) and Area Under Receiver Operator Characteristics (\textbf{AUROC}) are the two common threshold based metrics. In these evaluation metrics, area under the precision recall curve or ROC curve is measured by thresholding the posterior edge beliefs $q(\mG_{ij}\vert \mD)$ and averaging over all possible edges.
\end{itemize}
These metrics are easy to evaluate and have been widely used in prior works~\citep{lorch2021dibs,annadani2021variational,geffner2022deep,deleu2022bayesian,nishikawa2022bayesian,lorch2022amortized, atanackovic2023dyngfn}.  However, all these metrics evaluate samples from the posterior against a single graph (the ground truth) while ignoring the uncertainty due to finite data that makes other graphs plausible hypotheses.
\paragraph{Full posterior evaluation metrics.} The other metrics sample from the joint posterior over both $\mG$ and $\bphi$ to evaluate the goodness of the posterior approximation:
\begin{itemize}
    \item Negative Log-Likelihood: It is the negative Log-Likelihood (\textbf{NLL}) of held-out observational samples, computed by sampling the posterior model parameters, i.e. $-\E_{\mathbf{X}\sim p_\mathbf{X}(\mathbf{X})}\E_{q(\mG, \bphi\vert \mD)}\log p(\mathbf{X}\mid \mG, \bphi)$. Unlike in other inference problems like Variational Autoencoders~\citep{kingma2013auto,rezende2014stochastic}, NLL might not be the most suitable for structure learning because a graph with more edges has lower NLL than the ones with fewer edges.
    \item Interventional Negative Log-Likelihood: Since a posterior defines a generative model of data, interventional data of unseen interventions can be generated and compared with the ground truth data generative process. Interventional Negative Log-Likelihood (\textbf{I-NLL}) averaged over different unseen interventions is defined as: $-\frac{1}{d}\sum_{i=1}^d\E_{\mathbf{X}\sim p(\mathbf{X}\mid \mathrm{do}(X_i))}\E_{q(\mG, \bphi)}\log p(\mathbf{X}\mid \mG, \bphi, \mathrm{do}(X_i))$
    \item Interventional Distance Metrics: Similar to interventional negative log-likelihood, Interventional KL-Divergence (\textbf{I-KL}) and Interventional Maximum Mean Discrepancy (\textbf{I-MMD}) are metrics which measure the divergence between the unseen interventional distributions between the distribution induced by the generative model and that from the ground truth data generative process, i.e $\frac{1}{d}\sum_{i=1}^d\mathrm{D}(p_\mathbf{X}(\mathbf{X}\vert \mathrm{do}(X_i))\mid\mid q_\mathbf{X}(\mathbf{X}\vert \mathrm{do}(X_i))$ where $\mathrm{D}$ is either KL-divergence or maximum mean discrepancy~\citep{gretton2012kernel} and $q_\mathbf{X}$ is the data distribution induced by the approximated posterior.
\end{itemize}
NLL and I-NLL require that likelihood can be evaluated which might not be the case if the SCM is not an ANM. Given that most of the works deal with additive noise models, both these metrics have also been used in prior works~\citep{deleu2023joint,lorch2021dibs,annadani2023bayesdag,toth2022active,deleu2022bayesian, atanackovic2023dyngfn}.  


\begin{figure*}[t]
 \centering
 \includegraphics[width=0.8\textwidth]{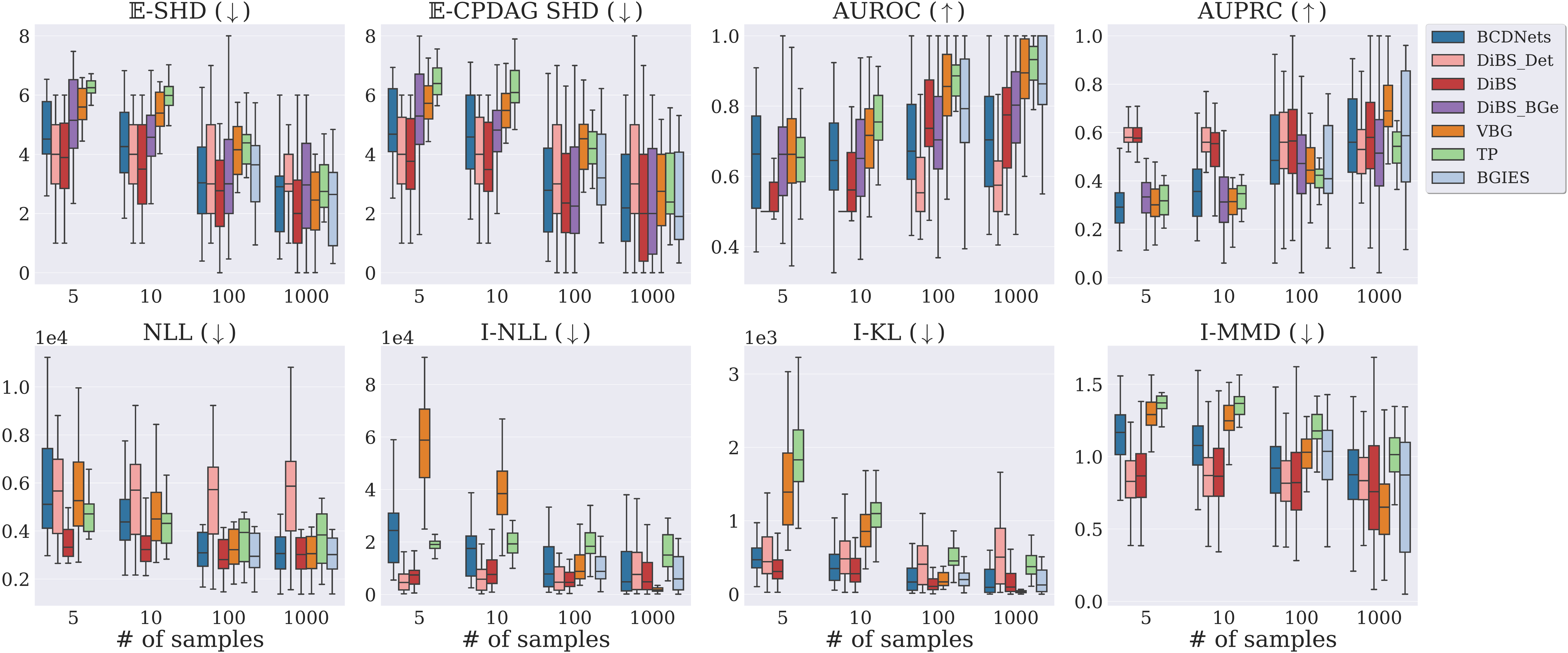}
 \caption{Evaluation of the models on ER1 graphs in the non-identifiable case ($d=5$). In low sample regimes, true posterior itself is evaluated to be worse on these metrics than their approximations.}
 \label{fig:er1-nonidentifiable}
\end{figure*}

\begin{figure*}[t]
 \centering
 \includegraphics[width=0.8\textwidth]{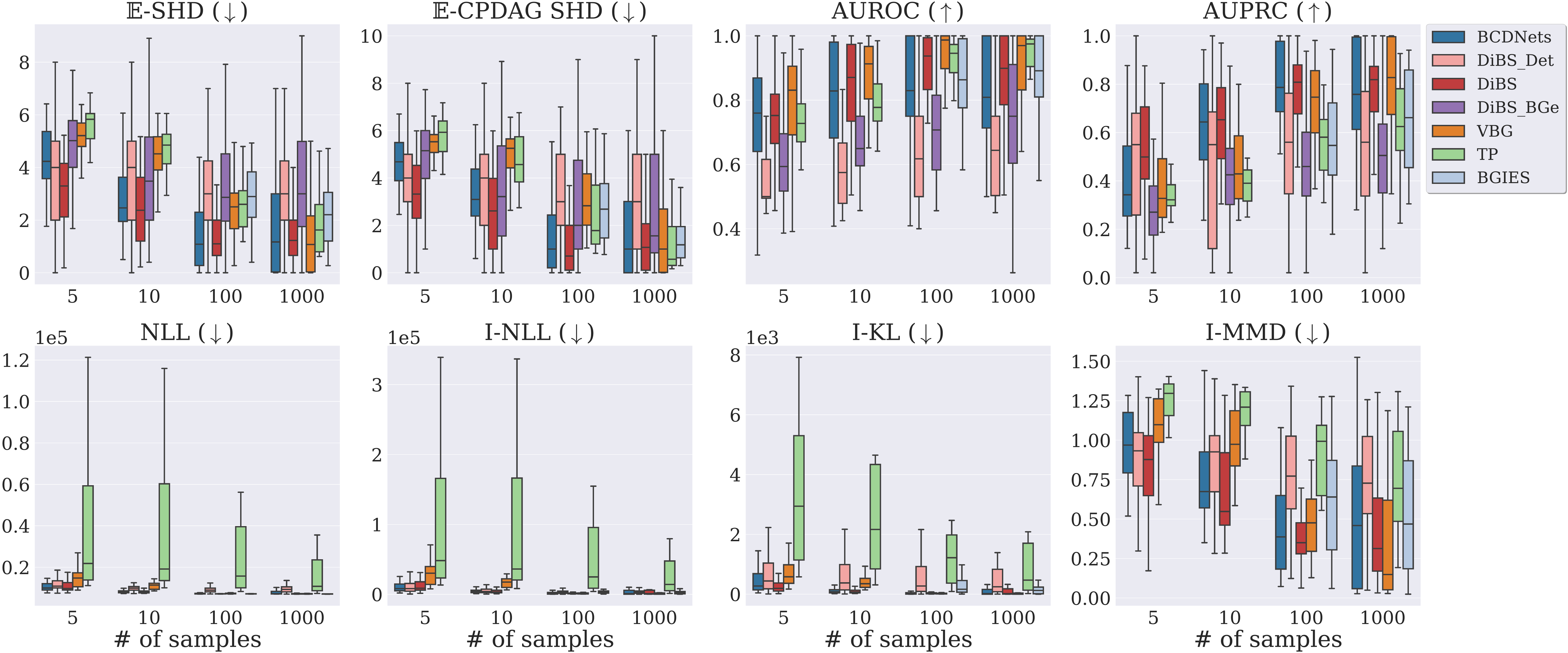}
 \caption{Evaluation of the models on ER1 graphs in the identifiable case ($d=5$). In low sample regimes, true posterior itself is evaluated to be worse on these metrics than their approximations.}
 \label{fig:er1-identifiable}
 \vskip -0.1in
\end{figure*}


\begin{figure*}[t]
 \centering
 \includegraphics[width=0.8\textwidth]{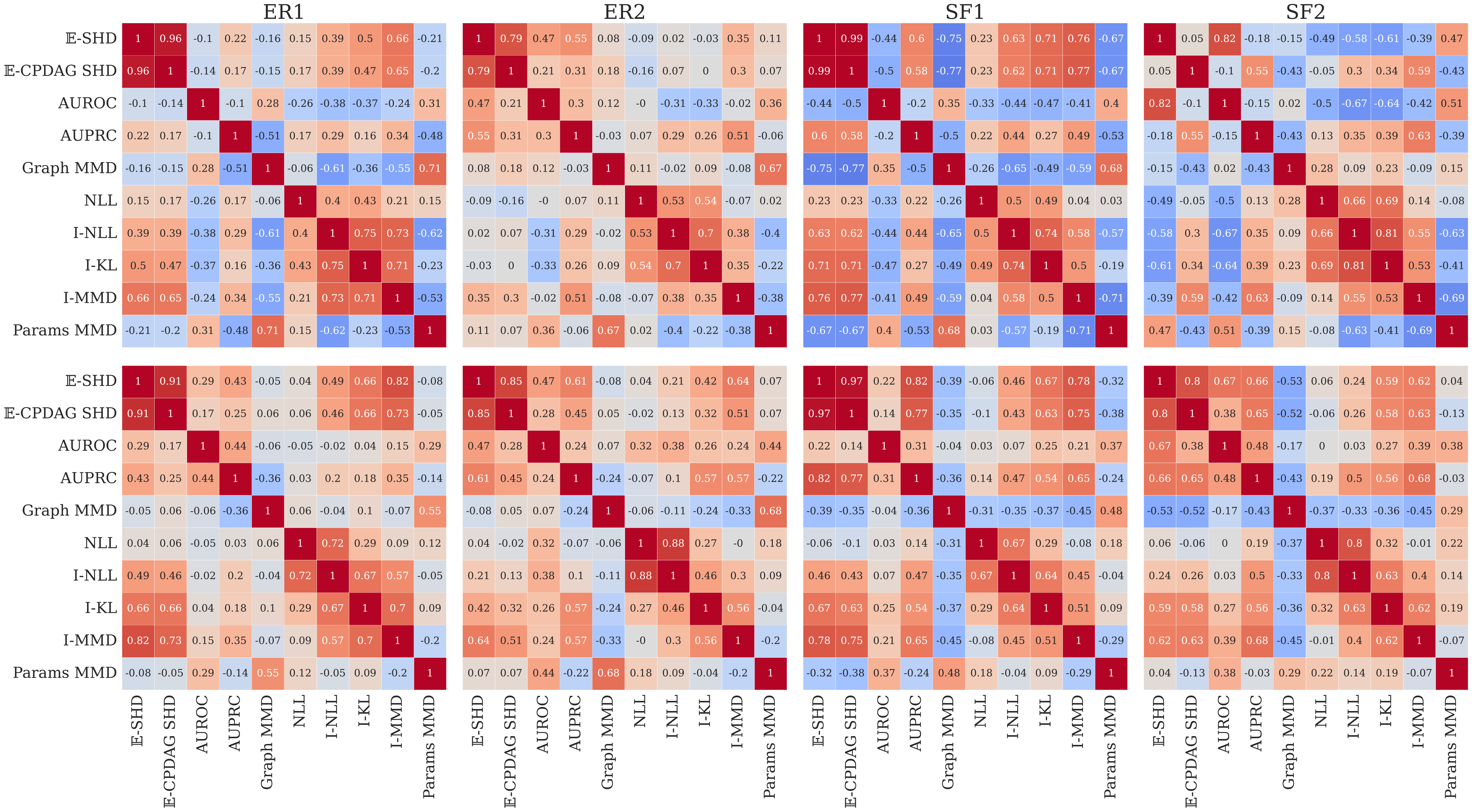}
 \caption{Spearman's rank correlation coefficient between evaluation metrics with 5 samples ($d=5$). The first and the second rows correspond to the non-identifiable and identifiable cases, respectively. All the graph-based metrics are not correlated with the Graph MMD. Params MMD is not correlated with any of the other metrics. Graph MMD and Params MMD are metrics that evaluate against the true posterior.}
 \label{fig:corr_5}
 \vskip -0.1in
\end{figure*}

\begin{figure*}[t]
 \centering
 \includegraphics[width=0.8\textwidth]{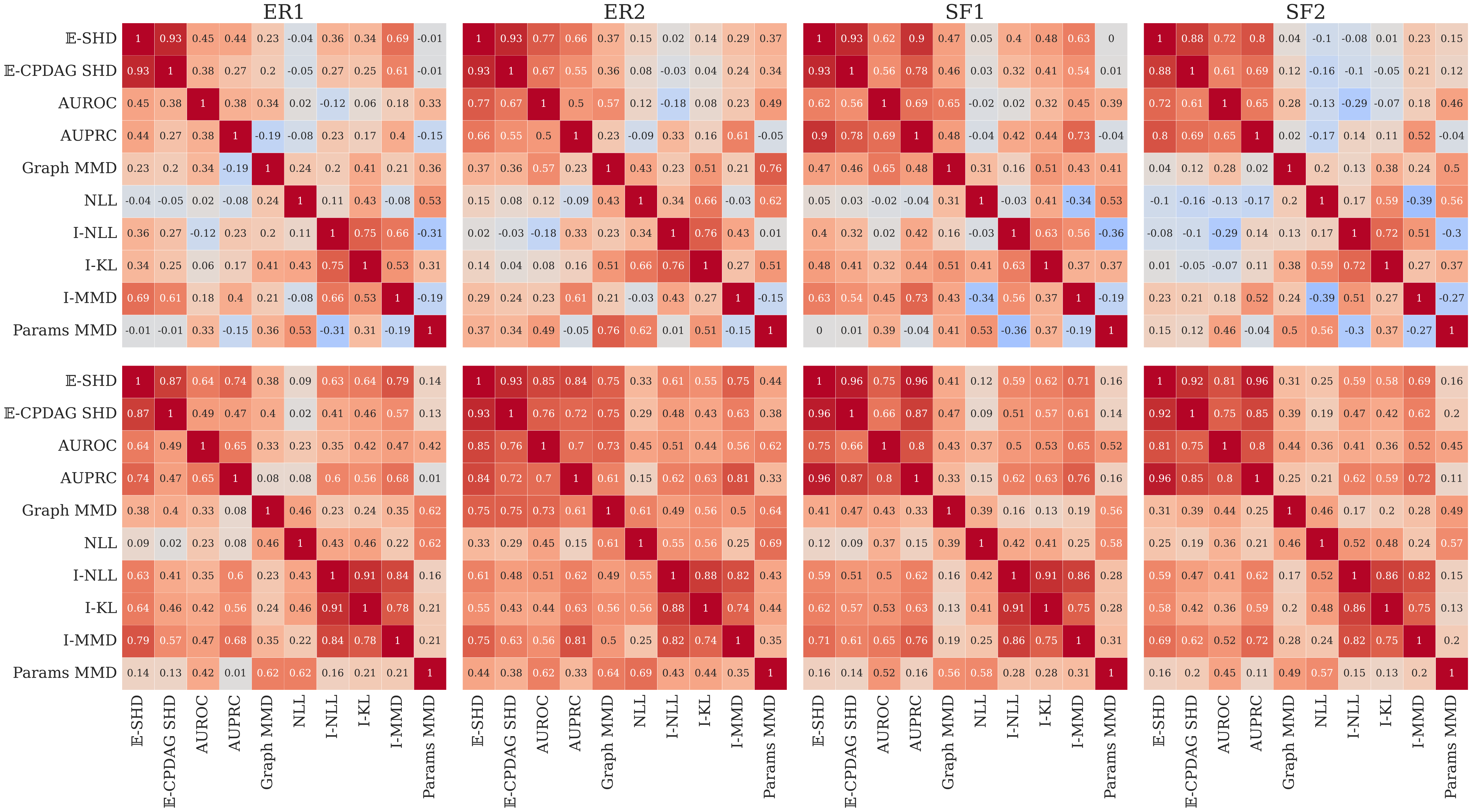}
 \caption{Spearman's rank correlation coefficient between evaluation metrics with 100 samples ($d=5$). The first and the second rows correspond to the non-identifiable and identifiable cases, respectively. All the graph-based metrics are correlated with each other and also the Graph MMD. Params MMD is also correlated with other metrics. Graph MMD and Params MMD are metrics that evaluate against the true posterior.}
 \label{fig:corr_100}
\end{figure*}

 Despite the extensive use of these metrics in prior work, it is unclear if they are suitable as proxy metrics. As BCD is a relatively new and emerging field, there is no principled case study yet which has addressed the evaluation problem. In the following section, we address this gap by performing an empirical study specifically with the aim to understand the evaluation metrics better. 
\section{Experiments and Key Results}
\label{sec:exp}
In this section, we design and perform a wide set of experiments on BCD methods to establish the suitability of the current evaluation metrics. We restrict our attention to linear additive noise models as most of the existing BCD methods are only applicable to this setting. In addition, true posterior can be computed in this setting in closed form, thus ensuring the drawn conclusions are sound. Model misspecification can be quite hard to deal with causal discovery in general~\cite{montagna2023assumption}. Therefore, we ensure in the experiments that all the methods have the same level of expressivity as the true posterior and have access to data with no model misspecification.
\paragraph{Outline of experiments.}
We mainly aim to understand the following aspects of the present evaluation metrics: (1) How does true posterior perform on these metrics? (2) Do all metrics correlate in terms of the ranking they induce on different models, and are they correlated with metrics which directly compare with the true posterior? (3) Entropy of the true posterior and how consideration of entropy of the true posterior is important for determining the reliability of the evaluation metrics (4) Downstream tasks that might be suitable for BCD when current metrics are not suitable.
\subsection{Experimental Setting}
\paragraph{Models.} We experiment on the following different BCD models: \textbf{BCD Nets}~\citep{cundy2021bcd}, \textbf{DIBS}~\citep{lorch2021dibs} and \textbf{VBG}~\citep{nishikawa2022bayesian} are methods which perform approximate inference on the graph, the parameters of the linear mechanisms and the variance of the noise variables. BCD Nets performs inference based on node permutation matrices using variational inference,  DIBS is a particle-based method with Stein Variational Gradient Descent (SVGD) \citep{liu2016stein} as its inference engine and VBG is a VI approach with GFlowNets~\citep{bengio2021flow}. We also include DAG  bootstrap~\citep{friedman1999data} with GIES~\citep{hauser2012characterization,chickering2002optimal} for evaluation though it is not strictly a Bayesian inference method as it has been used extensively as a baseline in prior work. Bootstrap GIES (\textbf{BGIES}) performs maximum-likelihood estimate on all the parameters of interest on different datasets bootstrapped from the original dataset, and then weighs each estimate with its unnormalized posterior probability. For comparison, we also include a non-Bayesian method by running DIBS deterministically (setting the number of particles of SVGD to 1), called  \textbf{DIBS Det}. Details of all the methods, including their hyperparameter search procedure are given in \cref{appendix:models}. When applicable, we also include a version of DIBS that directly uses the BGe score~\citep{geiger2002parameter} for likelihood (called \textbf{DIBS BGe}).
\paragraph{Synthetic data generation.} We test all the methods on synthetic data. This enables us access to ground truth as well as to have control over the SCM that generates the data, thereby ensuring there is no model misspecification. We sample graphs from Erdős–Rényi (ER)~\citep{erdHos1960evolution} and Scale-Free (SF)~\citep{barabasi1999emergence} random graph family along with a linear Gaussian ANM. The graphs have expected edge per node of either $1$  or $2$ (referred to as ER1, ER2, SF1 and SF2). We consider two scenarios for linear ANM: homoscedastic Gaussian (identifiable) and heteroscedastic Gaussian (non-identifiable)~\citep{peters2014identifiability}. In the first scenario, we
set the variance to one, while in the second scenario, the noise variances are sampled from an inverse gamma distribution. The weights are then derived from a multivariate Normal distribution with a mean of $0$ and a diagonal covariance matrix corresponding to noise variances. Details of the data generating process is given in \cref{appendix:data}. True posterior can be computed in closed form for both scenarios when $d<6$.  Details of true posterior computation is provided in \cref{appendix:tp_computation}. All the experiments are conducted with 20 different random datasets and 3 random initialization of the model per dataset, resulting in 60 runs for each model.

\paragraph{Metrics for comparison with true posterior.} As noted before, the true posterior is the gold standard with which the suitability of the other metrics can be reasonably established. In order to compare the approximate posterior with the true posterior, we use Maximum Mean Discrepancy (MMD)~\citep{gretton2012kernel} with relevant kernel. More precisely, we compare $q(\mG\vert \mD)$ with $p(\mG\vert\mD)$ (called \textbf{Graph MMD}) using a Hamming Kernel and $q(\bphi\vert \mG,\mD)$ with $p(\bphi\vert \mG, \mD)$ using an RBF kernel (called \textbf{Params MMD}). We use MMD as it requires only samples from the distribution.
\subsection{Key Results}

\begin{figure}[t]
 \centering
 \includegraphics[width=0.48\textwidth]{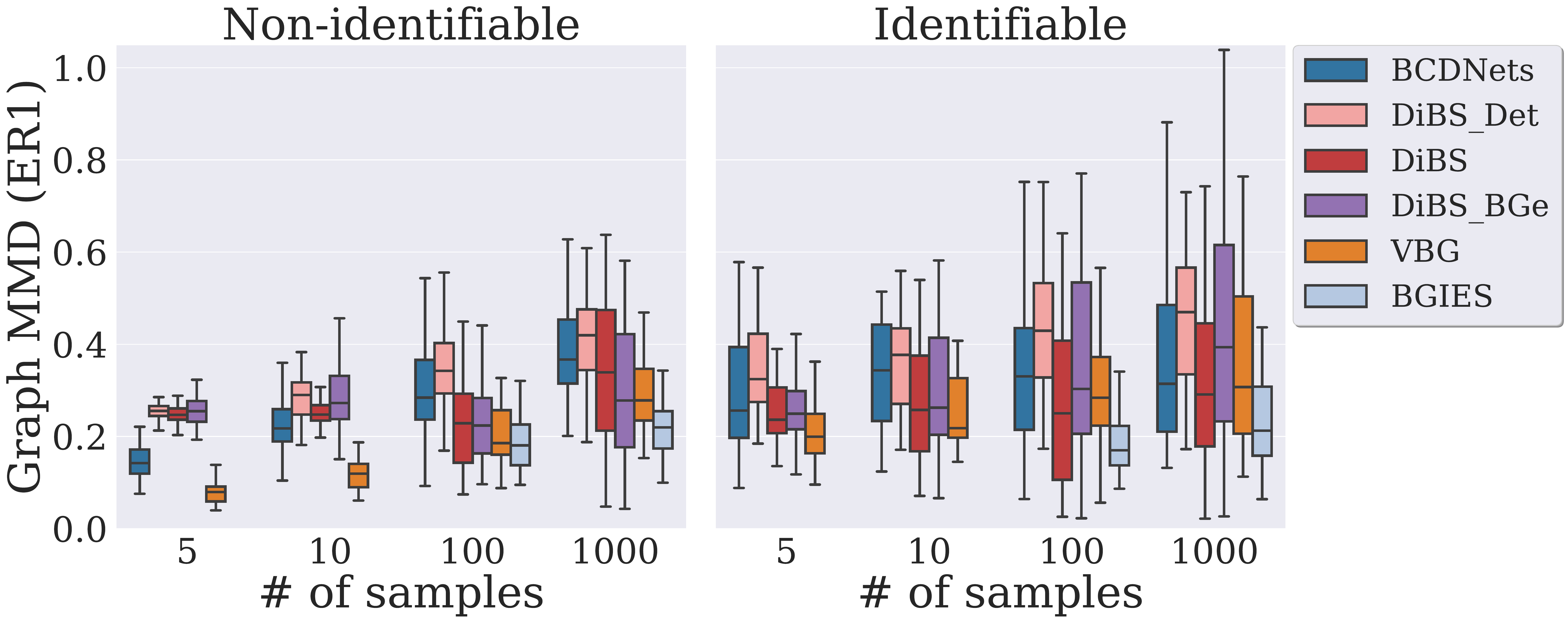}
 \vskip -0.05in
 \caption{Graph MMD of the models on ER1 graphs ($d=5$). }
 \label{fig:graph_mmd_er1}
\end{figure}

\begin{figure}[t]
 \centering
 \includegraphics[width=0.48\textwidth]{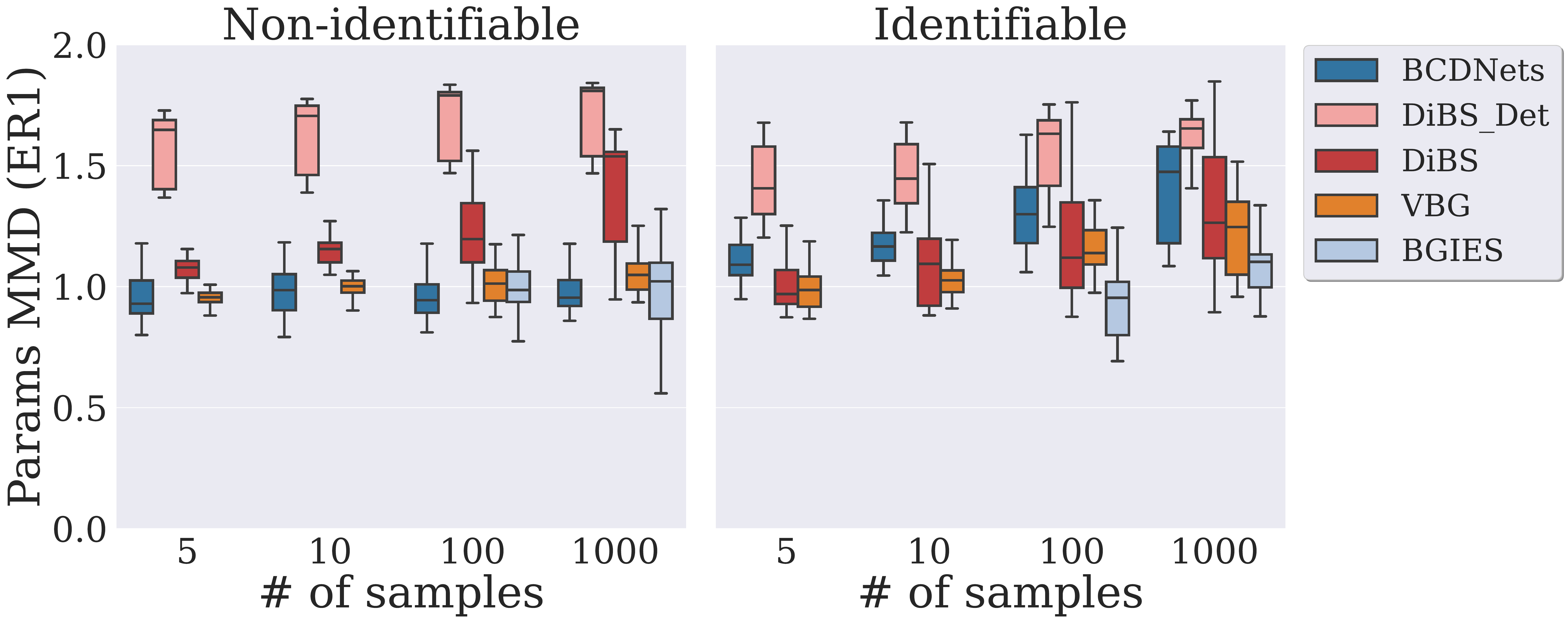}
 \vskip -0.05in
 \caption{Params MMD of the models on ER1 graphs ($d=5$).}
 \label{fig:params_mmd_er1}
 \vskip -0.1in
\end{figure}

\paragraph{Evaluation on current metrics.} We first evaluate all the methods on the metrics outlined in \cref{sec:metrics} to give a representative idea of the performance seen and reported in prior work. This would serve as a useful context for what is currently being evaluated in the literature. In addition, we also include the performance that true posterior achieves on these metrics. \cref{fig:er1-nonidentifiable} presents results for ER graph for non-identifiable setting and \cref{fig:er1-identifiable} for the identifiable setting. It is interesting to note that when the number of samples is smaller ($d=n=5$), the true posterior itself performs significantly worse on all metrics, including that of some of the BCD algorithms that are approximating the true posterior. For most applications, especially in biology, $n\approx d$ is a fairly common setting. In fact, many of the algorithms in BCD benchmark on synthetic datasets with the number of samples less than 100 (and in many cases just 50 samples, with $d$ ranging from 10 to 50). As the number of samples increases, the methods perform better on these metrics. However, the relative performance of all the methods, especially the true posterior, does not increase much when the number of samples is increased from 100 to 1000 (see \cref{appendix:example} for a simple example illustrating this point). This is consistent across different random graph models as well (\cref{Appendix:metrics_results}). As prior works mostly evaluate on higher dimensional cases where the true posterior is not tractable, this issue of worse performance of true posterior on these metrics has not been demonstrated before. At least preliminarily, this calls into question the suitability of the current evaluation metrics.
\paragraph{Evaluation on true posterior.} For comparison, we also present results that evaluate on true posterior with Graph MMD and Params MMD. \cref{fig:graph_mmd_er1,fig:params_mmd_er1} presents results for ER1 graphs. The evaluation indicates that the models considered do not estimate either the graph posterior or the parameter posterior well for $d=5$, as the MMD is greater than 0 for both cases.
Similar observation can be made for other graph types~(\cref{appendix:entropy_models}).


\paragraph{Rank correlation between metrics.} In order to further understand the suitability of current metrics, we analyze the Spearman's rank correlation coefficient between different metrics~\citep{spearman1961proof}. We rank different methods based on their performance in each metric and measure Spearman's rank correlation coefficient between rankings induced by each metric. It ranges between -1 to 1 -- a  coefficient of 1 would correspond to perfect correlation and -1 to inverse correlation. In other words, if Spearman's correlation between two metrics is -1, the model that is evaluated as the best under one metric would be evaluated by the other metric as the worst. With Spearman's rank correlation, we aim to analyze the following two questions: (1) Are the different proxy metrics correlated? and (2) More importantly, how correlated are the proxy metrics to the metrics that compare with the true posterior, i.e Graph MMD  and Params MMD? \cref{fig:corr_5} presents the result for non-identifiable scenario on a dataset with $d=5$, $n=5$. Several interesting conclusions can be drawn. Firstly, the graph-based proxy metrics are not correlated (for example $\E$-SHD and AUPRC), while the interventional-based metrics I-NLL, I-KL, and I-MMD are largely correlated. The correlation is higher in denser graphs. However, it is interesting to note that the interventional metrics do not correlate with NLL. Though the community has relied on NLL as a reasonable metric, it is sensitive to measurement errors and scale of the data~\citep{lorch2022amortized,reisach2021beware}. Secondly, all the graph-based metrics have very little to no correlation with graph MMD, and the Params MMD is not correlated with other metrics. 

In order to further understand if the same pattern exists in other settings, we examine the Spearman's rank correlation coefficient for the identifiable setting (\cref{fig:corr_5} bottom row). We observe a very interesting pattern. Unlike in the unidentifiable case, the graph-based metrics are more correlated, and the interventional metrics are correlated with each other and also with NLL. However, the graph-based proxy metrics are \textbf{not} well correlated with Graph MMD, although the level of correlation is slightly higher than the non-identifiable case. Similarly, Params  MMD is not correlated with other metrics.  This indicates that, while the metrics are usually correlated between each other in terms of ranking the models, the ranking that they induce would be different from the rankings induced by comparison with the true posterior when the number of samples is less.


\begin{figure*}[t]
 \centering
 \includegraphics[width=0.85\textwidth]{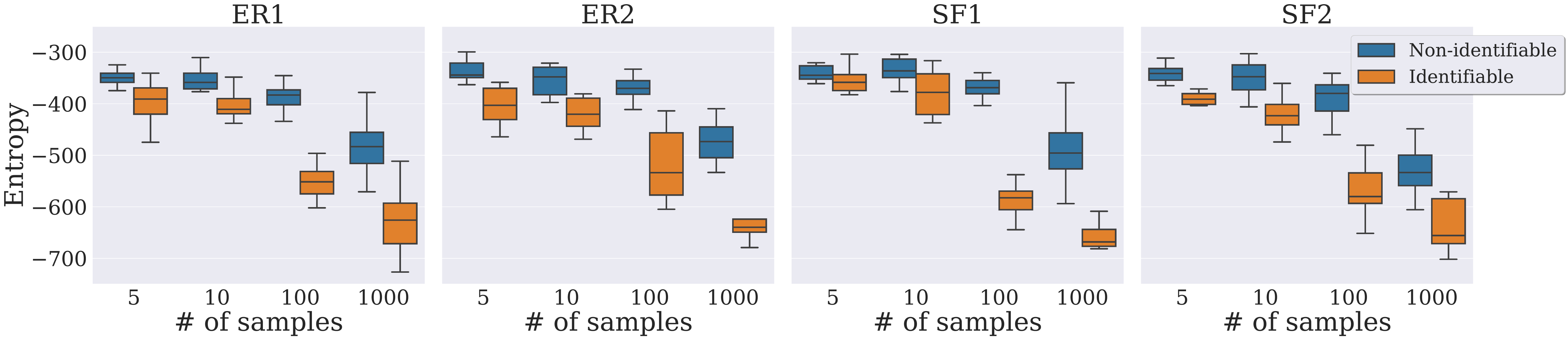}
 \caption{Entropy of the true posterior with different number of training samples ($d=5$). Entropy decreases as the number of samples increases. Entropy also decreases with identifiability.}
 \label{fig:entropy_tp}
\end{figure*}

\paragraph{Correlation between metrics for large datasets.} In order to see if the same correlation pattern persists for a higher number of samples, we plot Spearman's correlation coefficient for $N=100$ (\cref{fig:corr_100}). We observe that the correlation between Graph MMD and graph-based proxy metrics is higher than before, with the identifiable scenario having a much higher correlation than the non-identifiable one. A similar observation can be made for Params MMD. It is reasonable to expect based on this result that the current proxy metrics are viable for evaluation of BCD algorithms with more samples and an identifiable underlying SCM.

A similar observation when $N=\{10,1000\}$ (\cref{fig:corr_10,fig:corr_1000}) reveals that the proxy metrics are not correlated with the gold-standard metrics in practical settings with less data and non-identifiability, where being Bayesian about causal discovery is supposed to be advantageous. This calls into question the suitability of the current proxy metrics in these settings. 


\begin{figure}[t]
 \centering
 \includegraphics[width=0.48\textwidth]{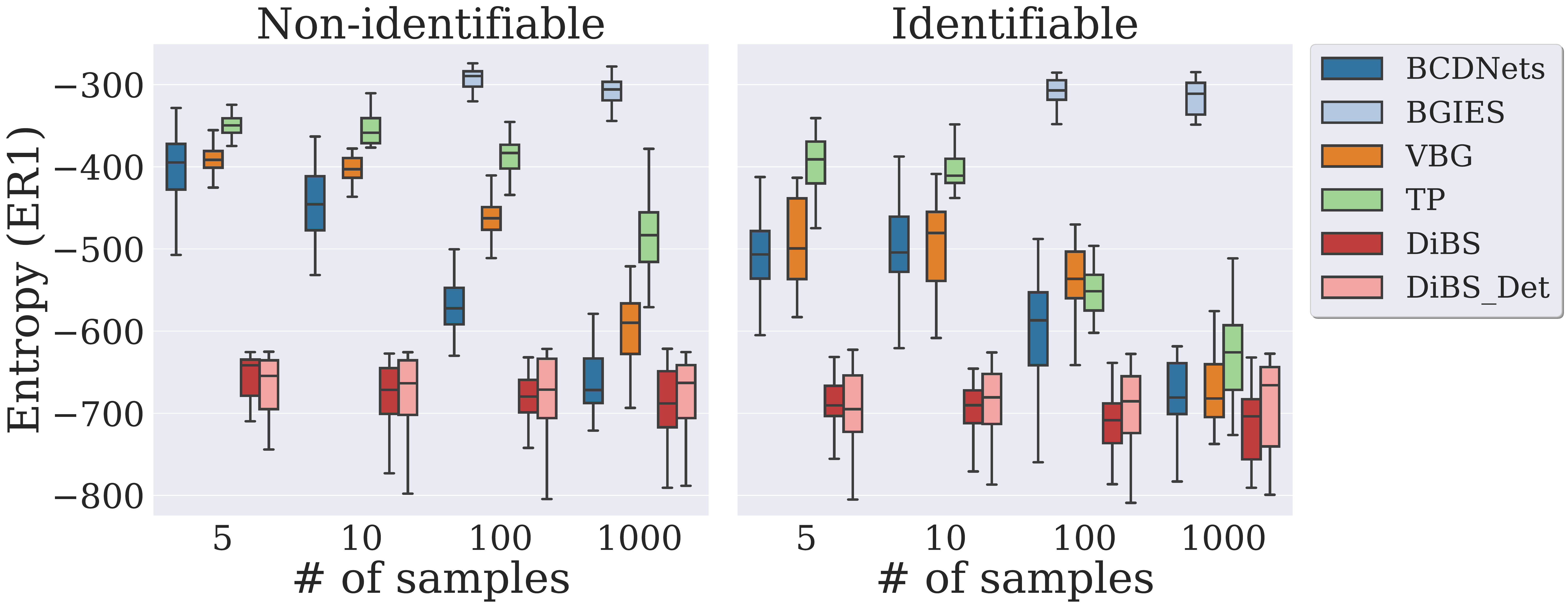}
 \vskip -0.05in
 \caption{Entropy of the models on ER1 graphs with different number of training samples ($d=5$). Entropy of most models except DIBS decreases with more data samples.}
 \label{fig:entropy_er1}
\end{figure}

\paragraph{Entropy of true posterior.} From the rank correlation, it is clear that the proxy metrics are only reliable with a high number of data samples and also depend on the nature of SCM, i.e. identifiability. We note that both of these aspects are connected to the entropy of the true posterior. In fact, it is reasonable to expect that the entropy of the true posterior decreases as the number of samples is increased. If an SCM is non-identifiable, all the graphs within the MEC,  which could be exponentially many, have a high probability, thereby making the posterior more entropic. We empirically demonstrate this on the true posterior corresponding to different settings. We use an approximator of entropy which only requires samples from the distribution~\citep{kozachenko1987sample}. Details are given in \cref{appendix:entropy_est}. \cref{fig:entropy_tp} illustrates the entropy of true posterior under various settings. It can be seen that entropy decreases with higher samples and identifiability. Since the proxy metrics are usually derived from causal discovery, they do not reflect the quality of approximation when there are many graphs (and corresponding parameters) with high posterior probability. Therefore, it is reasonable to conclude that the current metrics are not suitable where BCD is most desirable -- higher entropy settings of the true posterior.


\begin{figure}[t]
 \centering
 \includegraphics[width=0.31\textwidth]{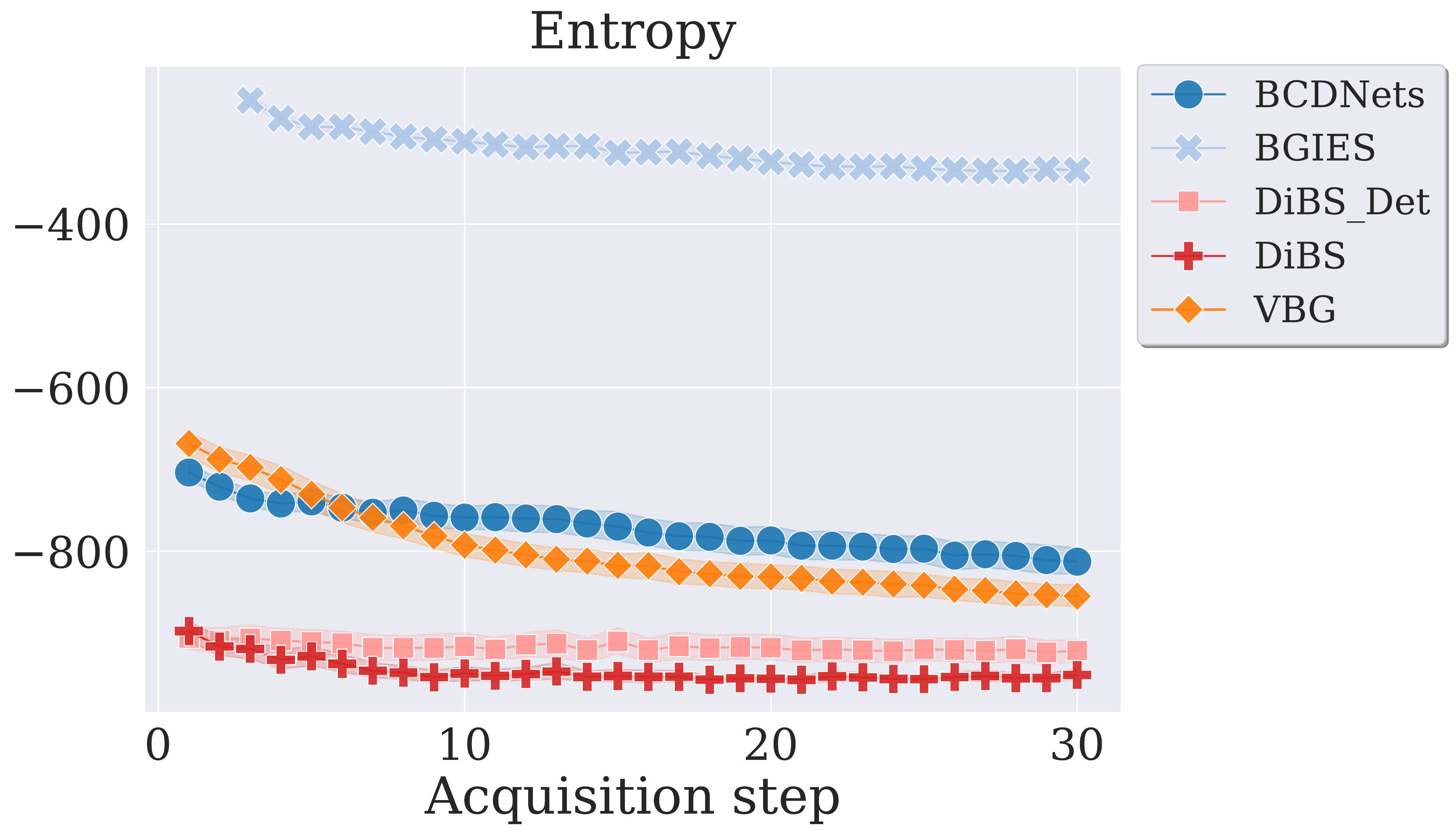}
 \caption{Changes of the entropy on ER1 graphs in an incremental setting in the non-identifiable case ($d=10$). We start with 10 observational samples and at each step, we add 5 interventional samples and retrain the models. For all the models, as we give them more interventional samples, the entropy does not decrease substantially.}
 \label{fig:incremental_entropy}
 \vskip -0.1in
\end{figure}

\paragraph{Entropy of models.} Using the same entropy estimator, we also examine the entropy of BCD models. In particular, we are interested in the following two aspects: 1) How entropic are the BCD algorithms in comparison to true posterior? 2) Does the entropy decrease as more observational and interventional data is given? Our goal is not to decide which method is the best but to understand if the methods respond to additional data to reduce their entropy. \cref{fig:entropy_er1} presents the results for ER1 graphs. Most of the methods have entropies the same as that of the true posterior, except DIBS, which always gives very low entropy solutions. Similar behaviour is seen in other graph types as well (\cref{appendix:entropy_models}). When interventional data is given and the model is updated at each step, similar to an experimental design loop~\citep{tigas2022interventions}, the reduction in entropy is very good for VBG and BCD Nets while it does not necessarily decrease for DIBS and BGIES (\cref{fig:incremental_entropy}). 

\begin{figure*}[t]
	\centering
	\includegraphics[width=0.8\textwidth]{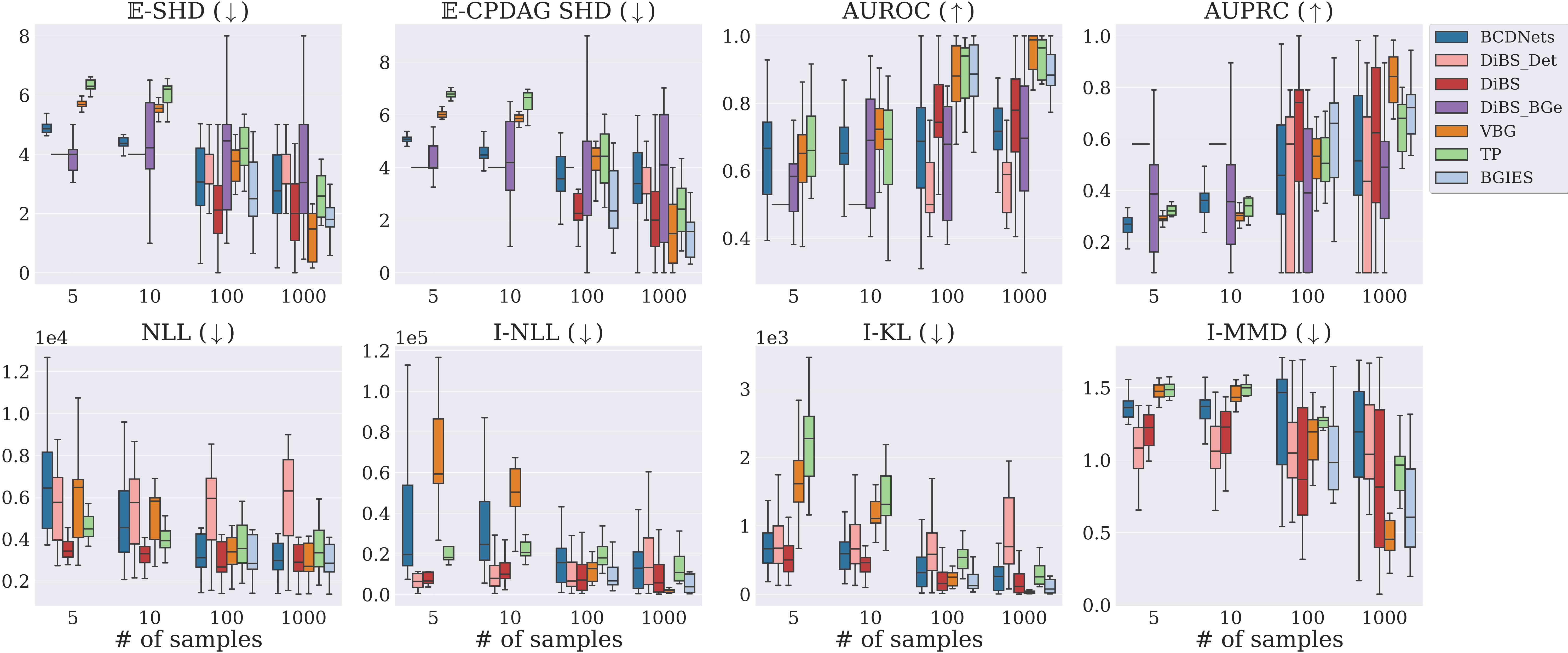}
	\caption{Evaluation of the models on SF1 graphs in the non-identifiable case ($d=5$). In low sample regimes, true posterior itself is evaluated to be worse on these metrics than their approximations.}
	\label{fig:sf1-nonidentifiable}
\end{figure*}

\paragraph{Effect of prior.} An important factor in obtaining a good estimate of the true posterior is the choice of the prior over the graphs and parameters of the model, i.e. $p(\mG, \bphi)$. Apart from DIBS, all methods do not use an informative prior.  DIBS leverages the knowledge of the underlying data generative process to design priors that match that are informative. While a more extensive study is required to understand the performance of various methods due to the choice of the prior, we do notice that for SF graphs, the solution of DIBS is completely dominated by the prior in low data regimes. While this is the intended behavior in a Bayesian setting, the solution of DIBS is very low entropy. In fact, we found that it samples a graph with no edges in low data regimes (\cref{fig:sf1-nonidentifiable,fig:sf2-nonidentifiable}). However, for ER graphs, the prior is less dominant than for SF graphs and it leads to reasonable samples with DIBS.

\section{Possible Alternative Evaluation Procedures}
\label{sec:alternative_evaluation}
Although our study mainly focuses on identifying the potential issues in the evaluation metrics for BCD, we suggest two possible alternate way of evaluating BCD algorithms by considering the empirical results obtained in \cref{sec:exp}.
\subsection{Experimental Design}
As seen in \cref{fig:entropy_tp}, after acquiring enough (interventional) data, the true posterior will have less entropy. Therefore, one possible way to evaluate the BCD algorithms is to evaluate it downstream, for instance, by performing experimental design to acquire enough interventional data and then evaluating with the proxy metrics when they are more suitable. The task of choosing the intervention that results in the highest expected reduction in entropy is concerned with  Bayesian optimal experimental design~\citep{lindley1956measure,chaloner1995bayesian,foster2019variational}, a downstream task of Bayesian Causal Discovery. Many specific experimental design procedures exist for BCD~\citep{tigas2022interventions,tigas2023differentiable,agrawal2019abcd,zhang2023bayesian,toth2022active} that can be used to collect data and perform evaluation.   

\subsection{Causal Effect Estimation}

In some applications, proxy metrics either require access to the underlying ground truth graph or other parameters thereof, which might not be available. In such a case, experimental design as a downstream evaluation tool might not be applicable.  
An alternative evaluation procedure in such a case, therefore, is the downstream task of causal effect estimation. Causal effect estimation is the task of estimating the state of a variable that is part of the causal model when the system is subject to interventions. This method has been thoroughly studied in \citet{emezue2023benchmarking} and has shown to be useful in identifiable cases with few data samples.


\section{Discussion and Conclusion}
\label{sec:conclusion}
In this work, we demonstrate the shortcomings of the present evaluation metrics for Bayesian Causal Discovery with an extensive empirical study. Our key result is that when the true posterior has high entropy - which is the case with less data and non-identifiability, current evaluation metrics do not lead to the same ranking of different BCD models compared to metrics that involve the true posterior. Therefore, evaluation of BCD should be considered carefully in settings with limited data and identifiability of SCM. This challenge of evaluating BCD in these settings could potentially be overcome by evaluation in downstream tasks: for example, causal effect estimation or  Bayesian experimental design to acquire interventional samples, after which the true posterior has less entropy which enables reliable evaluation with current metrics.
While our study sheds a light on evaluation procedures and their shortcomings in BCD, our study is limited to causally sufficient linear additive noise models. As the field of Bayesian Causal Discovery progresses in terms of posterior approximation in settings where these assumptions do not hold, a similar analysis as presented in this work might be necessary for such settings.

\section*{Acknowledgements}
This work was partially supported by the Wallenberg AI, Autonomous Systems and Software Program (WASP) funded by Knut and Alice Wallenberg Foundation, and the computations were enabled by the Berzelius resource provided by the Knut and Alice Wallenberg Foundation at the National Supercomputer Centre.

\section*{Impact Statement}
This work is concerned with proper evaluation of Bayesian causal discovery algorithms which highly benefits the research community. The authors do not foresee negative societal impact of this work beyond what is brought about by general advances in machine learning.



\bibliography{example_paper}
\bibliographystyle{icml2024}

\newpage
\appendix
\onecolumn
\section{Experimental Details}
\subsection{Models}
\label{appendix:models}
This section provides a casual overview of the models featured in the study, along with details regarding their implementation and the choices made for hyperparameters.

\paragraph{BCD-Nets.} BCD-Nets \citep{cundy2021bcd} is a bayesian posterior approximation method designed to model linear-Gaussian SCMs. It decomposes the weighted adjacency matrix $W$ of the linear SCM into a permutation matrix $P$ and a strictly lower-triangular matrix $L$, i.e. $W = PLP^T$. It uses variational inference to learn the posterior distribution $q_\phi(P, L, \Sigma)$ over the SCM parameters 
by maximizing the Evidence Lower Bound (ELBO) w.r.t. variational parameters $\phi$. For the implementation, we utilize the public implementation of BCD-Nets \footnote[1]{\href{https://github.com/ermongroup/BCD-Nets}{https://github.com/ermongroup/BCD-Nets}} with the same hyperparameters as in the original paper except for the number of training steps which we change to $20$k steps.

\paragraph{DIBS.} DIBS \citep{lorch2021dibs} is a fully differentiable bayesian posterior approximation method suitable to model both linear and non-linear SCMs. It proposes to transfer the posterior inference into the latent space of a probabilistic graph representation and assumes there is a latent variable $Z$ that models the generative process of the underlying causal graph. They factorize the joint distribution $p(Z, G, \Theta, D)$ in a way that allows for joint posterior inference of both the graph structure and the conditional distribution parameters. 
To be more precise, $p(Z, \mG, \bphi, \mD) = p(Z)p(\mG \mid Z)p(\bphi \mid \mG)p(\mD \mid \mG, \bphi)$.
They apply the gradient-based SVGD algorithm \citep{liu2016stein} for sampling. In this work, we utilize $3$ different versions of DiBS+: the linear version (we refer to as \textit{DIBS}), a deterministic variant of \textit{DiBS} in which we have only $1$ particle in the model (referred to as \textit{DIBS-Det}), and a marginal version (we refer to as \textit{DIBS-Bge}) where the marginal posterior over the graphs, i.e. $p(G|D)$, is computed using the Bayesian Gaussian Equivalent (\textit{BGe}) marginal likelihood. Also, we use the implementation of \citet{tigas2022interventions}\footnote[2]{\label{cbed}\href{https://github.com/yannadani/cbed}{https://github.com/yannadani/cbed}} and change it to learn the noise variances together with other parameters. For all experiments, we set the $\sigma_z$, $\alpha$, $\gamma_z$, and $\gamma_\theta$ to $0.5$, $0.02$, $5$, and $500$, respectively, use $50$ particles, and run the model for $20$k iterations. We use the default values for other hyperparameters.

\paragraph{VBG.} VBG \citep{nishikawa2022bayesian} is another bayesian posterior approximation model suitable designed to model linear-Gaussian SCMs. It extends the \textit{DAG-GFlowNet} \citep{deleu2022bayesian} to not only learn the graph structure but also the parameters of a linear Gaussian model between the variables in the DAG. In order to model the posterior distribution over the parameters, it utilizes \textit{GFlowNets} \citep{bengio2021flow}. We use the public implementation of VBG \footnote[3]{\href{https://github.com/mizunt1/vbg}{https://github.com/mizunt1/vbg}} and use its default hyperparameters for all experiments. As VBG assumes fixed noise variances, we experimented with various values for the noise variance and determined that $0.1$ yields the best results in our settings.

\paragraph{DAG Bootstrap.} DAG Bootstrap \citep{friedman1999data} is a non-parametric model that performs model averaging by bootstrapping the data to yield a collection of synthetic datasets. Each dataset is then utilized to learn an individual graph and its associated causal mechanisms, employing the score-based GIES algorithm \citep{chickering2002optimal,hauser2012characterization}. The ensemble of distinct single graphs approximates the posterior by assigning weights to each graph based on its unnormalized posterior probability. For our experiments, we employ the implementation of \citet{tigas2022interventions}\footref{cbed}, and use $100$ bootstraps.
\subsection{Synethetic Dataset Details}
\label{appendix:data}
In this study, we adopt Erdős–Rényi (ER)~\citep{erdHos1960evolution} and Scale-Free (SF)~\citep{barabasi1999emergence} graphs as the underlying graph structures for all experiments, utilizing a linear structural equation model (SEM) with $5$ and $10$ nodes. We generated the graph by setting an expected edge of $1$ or $2$ for each node. For the SCM weights and noises, we considered two scenarios. In the first scenario, we introduce Gaussian noise with equal variances across all nodes, with the variance value set to $1$, and sample the weights of the SCM from independent normal distributions with the mean and the variance set to $0$ and $2$ respectively. In this scenario, the underlying causal model will be identifiable. In the second scenario, we explore a non-equal variance case where the noise variances are sampled from an inverse gamma distribution with $\alpha$ and $\beta$ set to $4$ and $0.5$, respectively. The parameters of the inverse gamma distribution are chosen to restrict the noise variances to a low value, preventing the generation of data with high levels of noise. The weights are then derived from a Multivariate Normal distribution with a mean of $0$ and a diagonal covariance matrix corresponding to noise variances. Data was then sampled using ancestral sampling, and different numbers of training samples ($N=\{5,10,100,1000\}$ were generated for different experiments.
\section{Limitations of Graph Metrics: Example}
\label{appendix:example}
\begin{figure}[h]
	\centering
	\includegraphics[width=0.8\textwidth]{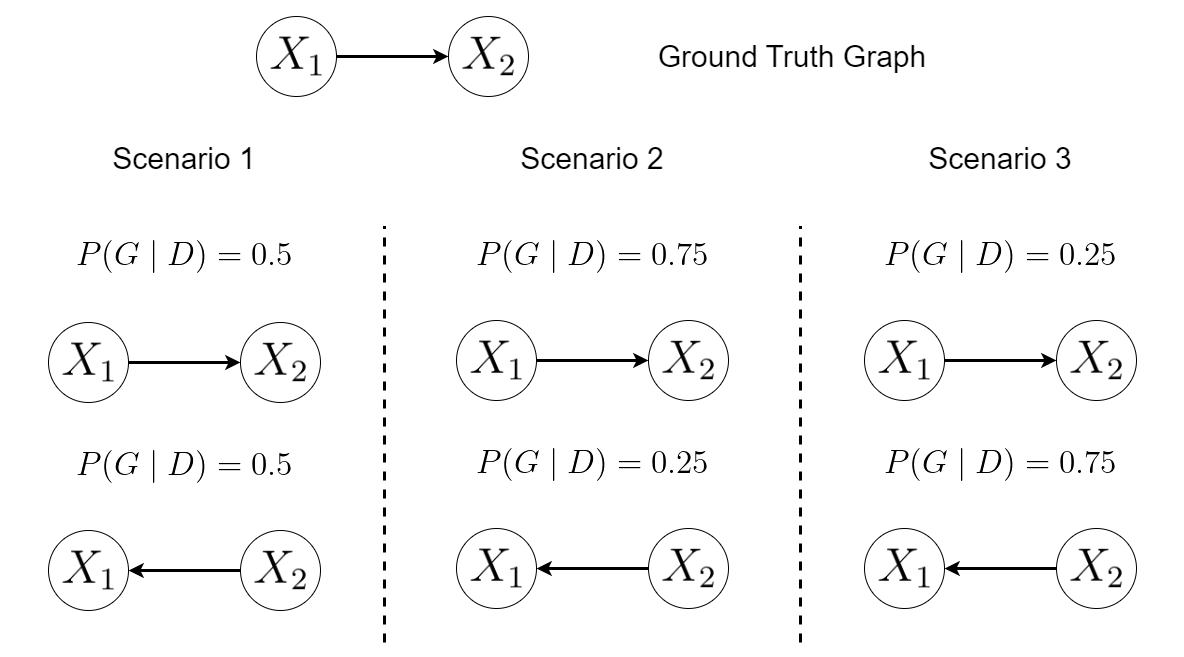}
	\caption{A simple example showing the shortcomings of the graph-based metrics in evaluating posterior distributions. Each scenario corresponds to a posterior distribution over the possible graphs. In the first scenario, $\E$-SHD is $0.5$, $\E$-CPDAG SHD is $0$, AUROC is $0.66$, and AUPRC is $0.5375$. In the second scenario, $\E$-SHD is $0.25$, $\E$-CPDAG SHD is $0$, AUROC is $1$, and AUPRC is $0.775$. In the third scenario, $\E$-SHD is $0.75$, $\E$-CPDAG SHD is $0$, AUROC is $0.66$, and AUPRC is $0.2625$. Suppose if the model was non-identifiable, then our posterior would correspond to scenario $1$ even with lots of samples. However, we don't necessarily get the best performance as evaluated by these metrics. Rather, approximate inference method might lead to solutions similar to other scenarios (for example, scenario $2$) which are evaluated to be better than the true posterior (scenario one).}
	\label{fig:example}
\end{figure}
\section{True Posterior Computation}
\label{appendix:tp_computation}
Note that for an ANM, the likelihood can be evaluated through the noise variable, which we assume to be Gaussian~\citep{geffner2022deep}. Therefore, $p(\mD\mid \mG, \bphi) = \prod_{j=1}^N\prod_{i=1}^d\mathcal{N}(\bgamma_i^T X_{\text{pa}(i)}^{(j)},\sigma^2_i)$
\subsection{Parameter Posterior}
We follow the posterior computation from~\citep{cho2016reconstructing}. More precisely, $\sigma^2\sim \text{Inv-Gamma}(\alpha,\beta)$ and $\bphi_i\sim \mathcal{N}(\mu_i, \sigma^2(\Lambda_i)^{-1})$. Let $\mathbf{X}_{\text{pa}(i)}\in \mathbb{R}^{N\times \vert \text{pa}\vert}$ be the matrix of parents for variable $i$ and $\mathbf{X}_i\in \mathbb{R}^N$ be the vector of samples of variable $i$. The posterior over parameters has the same form with parameters for a given graph:
\begin{align*}
    \Lambda'_i &\coloneqq \mathbf{X}_{\text{pa}(i)}^T\mathbf{X}_{\text{pa}(i)} + \Lambda_i\\
    \mu'_i &\coloneqq (\Lambda'_i)^{-1}(\Lambda_i\mu_i + \mathbf{X}_{\text{pa}(i)} \mathbf{X}_i)\\
    \alpha' &\coloneqq \alpha + \frac{N}{2}\\
    \beta' &\coloneqq \beta + \frac{1}{2}(\mathbf{X}_i^T \mathbf{X}_i + \mu_i^T\Lambda_i \mu_i - (\mu'_i)^T\Lambda'_i \mu'_i)
\end{align*}
In this work, we set $\alpha=4$, $\beta=0.5$ and $\Lambda = \mathbb{I}$.
\subsection{Graph Posterior}
The marginal likelihood function $p(\mD\mid \mG)$ can also be obtained in closed form, through which the graph posterior $p(\mG \mid \mD)$ can be derived by enumerating all the possible graphs.
For the identifiable case, the marginal likelihood is given by:
\begin{equation*}
    p(\mD\mid \mG) = (2\pi)^{Nd}\cdot \frac{(\beta)^{d\alpha}}{(\beta')^{d\alpha'}}\cdot \frac{\Gamma(\alpha')^d}{\Gamma(\alpha)^d} \prod_{i=1}^d \sqrt{\frac{\mathrm{det}(\Lambda_i)}{\mathrm{det}(\Lambda'_i)}} 
\end{equation*}

If the posterior has to ensure that all the graphs within the MEC have the same probability given large number of samples, it can be ensured with the BGe score~\citep{geiger2002parameter}. The marginal likelihood is given in ~\citep{kuipers2014addendum} (Equation 6), and we use the implementation of \citep{lorch2021dibs}. Note that BGe score assumes that the parameter priors are sampled from a Gaussian-Wishart distribution, instead of Gaussian-Inverse Gamma. Although strictly this assumption is violated in our data generative process, the computation of $p(\mG|\mD)$ is still valid. 
\section{Entropy Estimator}
\label{appendix:entropy_est}
For any random variable $\mathbf{Y}\in \mathbb{R}^p$, the Kozachenko-Leonenko estimate of the entropy $\mathrm{H}(\mathbf{Y})$, with N iid samples from $p_\mathbf{Y}$ is given by~\citep{kozachenko1987sample}:
\begin{equation}
    \hat{\mathrm{H}}_{\text{KL}}(\mathbf{Y}) = \psi(N) - \psi(n) + \log(c_p) + \frac{p}{N}\sum_{i=1}^N \log (\epsilon(i))
\end{equation}
where $\epsilon(i)$ is the distance of the $i$\textsuperscript{th} sample to its $n$\textsuperscript{th} nearest neighbor, $c_p = \frac{\pi^{\frac{p}{2}}}{\Gamma(1+\frac{p}{2})}$, $\psi(\cdot)$ is the digamma function and $\Gamma(\cdot)$ is the Gamma function.
As $\mathbf{Y}$ corresponds to parameters of the causal model and the causal graph in our case, we measure $\hat{\mathrm{H}}_{\text{KL}}(\mathbf{Y})$ of the distance between likelihoods of the samples induced by the posterior estimates, as that would reflect the information geometry of the approximate posterior better than the parameters themselves.
More precisely, we measure the Kozachenko-Leonenko estimate of the entropy on between distances of likelihoods of held-out data as measured by a kernel.
\begin{equation}
    \hat{\mathrm{H}}(\mG,\bphi) \approx \hat{\mathrm{H}}_{\text{KL}}\left[\mathop{\E}_{p_\mathbf{X}}\mathop{\E}_{\mG',\bphi'\sim q}\left[k(\log p(\mathbf{X}\mid \mG,\bphi), \log p(\mathbf{X}\mid \mG',\bphi'))\right]\right]
\end{equation}
where $k(\cdot,\cdot)$ is the RBF kernel. We use the implementation provided by \citep{lombardi2016nonparametric}.
\newpage

\section{Additional Results}
\label{appendix:plots}

In this section, we report additional results and show the evaluation of models on different graph types.

\subsection{Effect of Data Normalization.} 
\begin{figure}[h]
	\centering
	\includegraphics[width=0.5\textwidth]{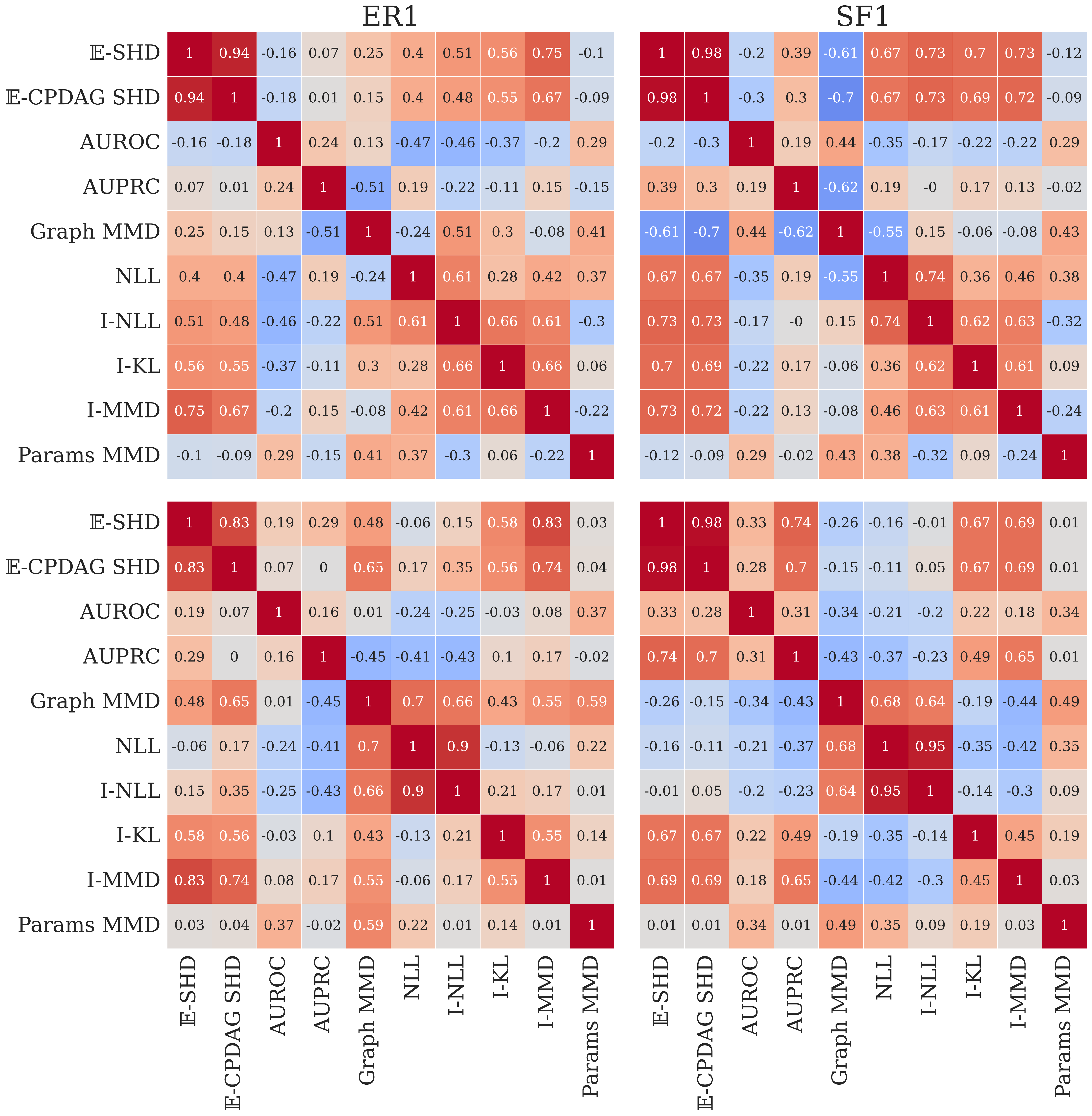}
	\caption{Spearman's rank correlation coefficient between evaluation metrics with 5 normalized training samples ($d=5$). The first and the second rows correspond to the non-identifiable and identifiable cases, respectively. Similar to the unnormalized case, all the graph-based metrics are not correlated with the Graph MMD, and Params MMD is not correlated with any of the other metrics. Graph MMD and Params MMD are metrics that evaluate against the true posterior.}
	\label{fig:corr_normalized_5}
\end{figure}

\begin{figure}[h]
	\centering
	\includegraphics[width=0.5\textwidth]{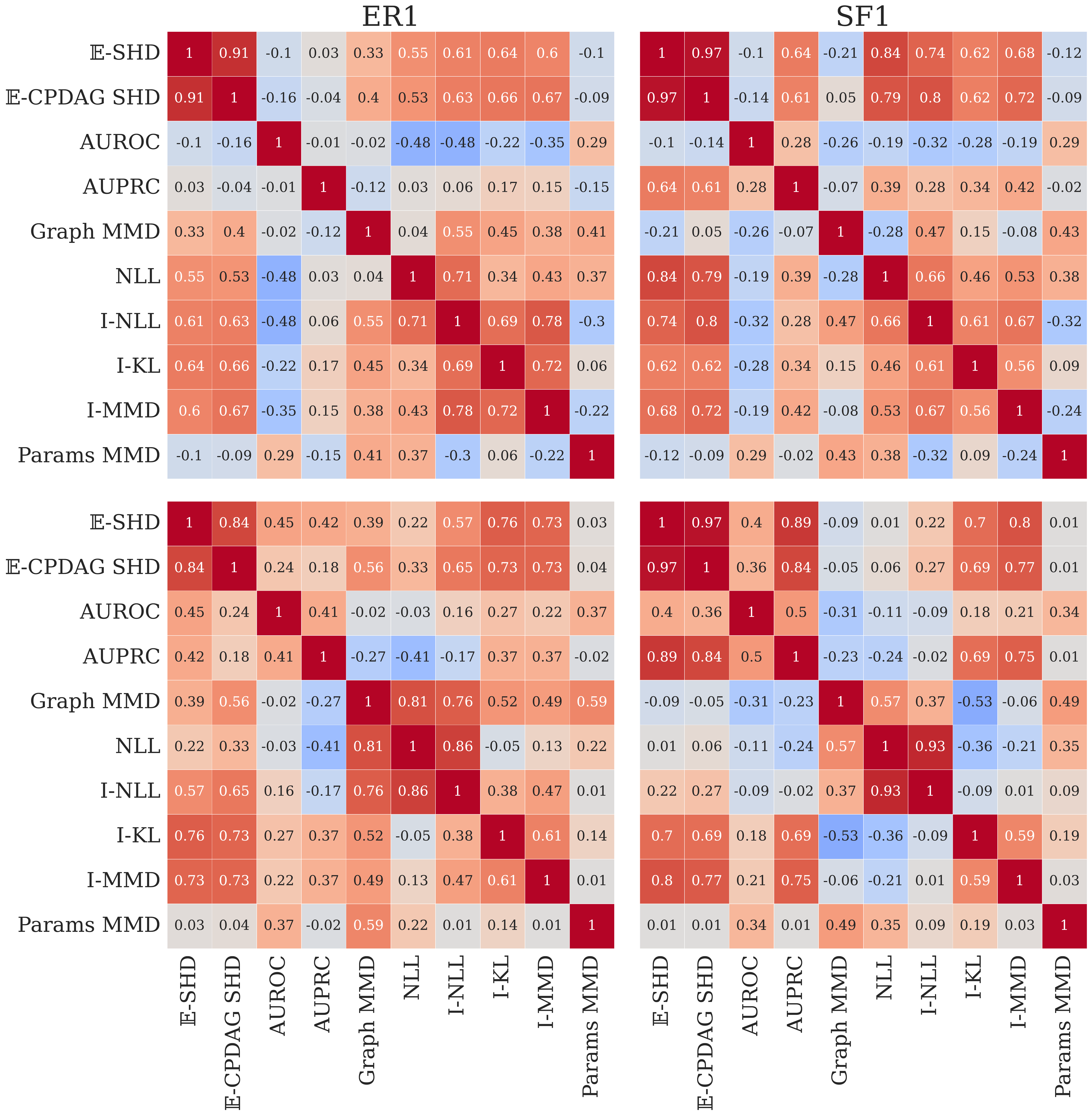}
	\caption{Spearman's rank correlation coefficient between evaluation metrics with 10 normalized training samples ($d=5$). The first and the second rows correspond to the non-identifiable and identifiable cases, respectively. The same pattern as in the non-normalized case is observed.}
	\label{fig:corr_normalized_10}
\end{figure}

\begin{figure}[h]
	\centering
	\includegraphics[width=0.5\textwidth]{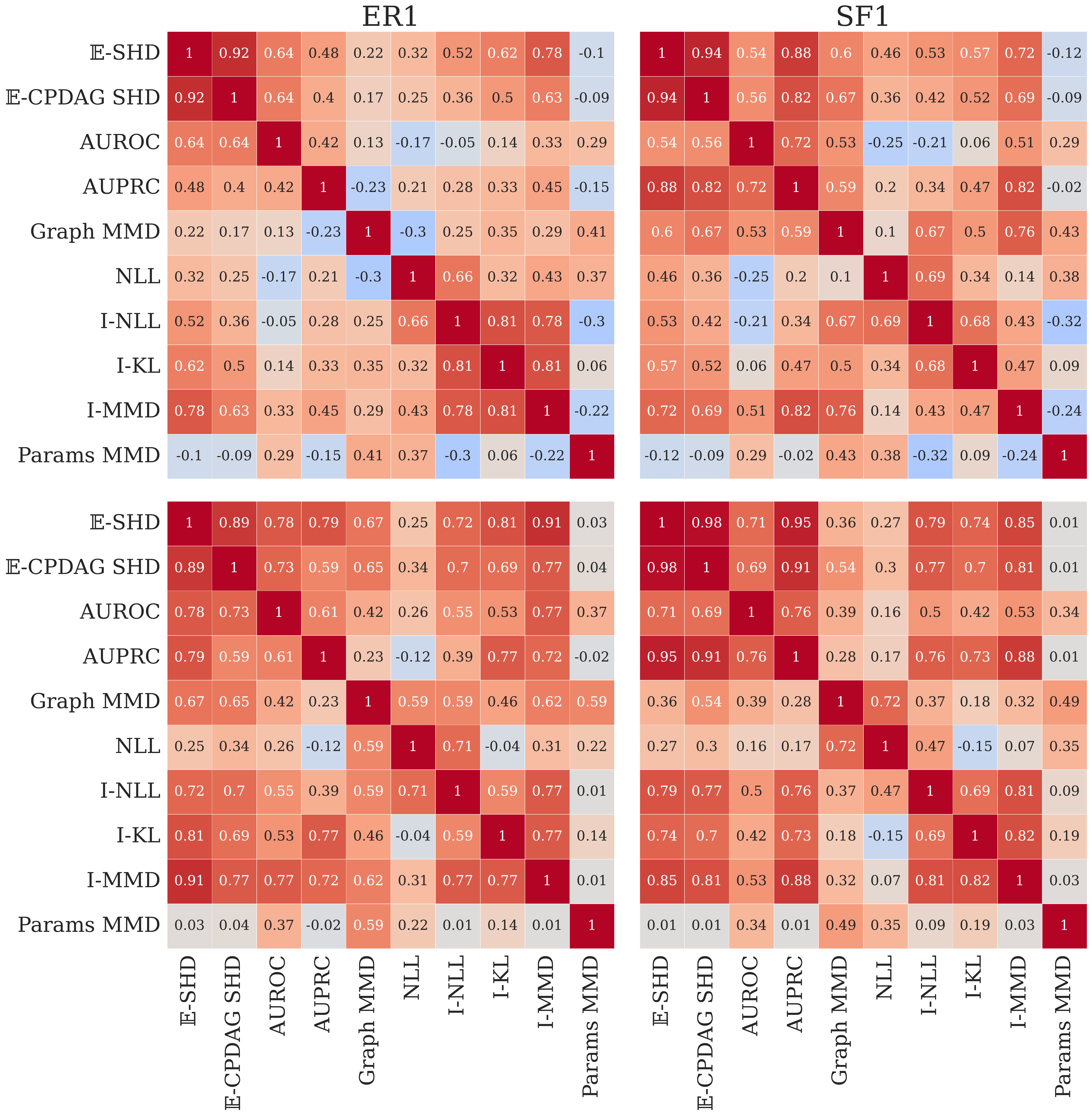}
	\caption{Spearman's rank correlation coefficient between evaluation metrics with 100 normalized training samples ($d=5$). The first and the second rows correspond to the non-identifiable and identifiable cases, respectively. The same pattern as in the non-normalized case is observed.}
	\label{fig:corr_normalized_100}
\end{figure}

\begin{figure}[h]
	\centering
	\includegraphics[width=0.5\textwidth]{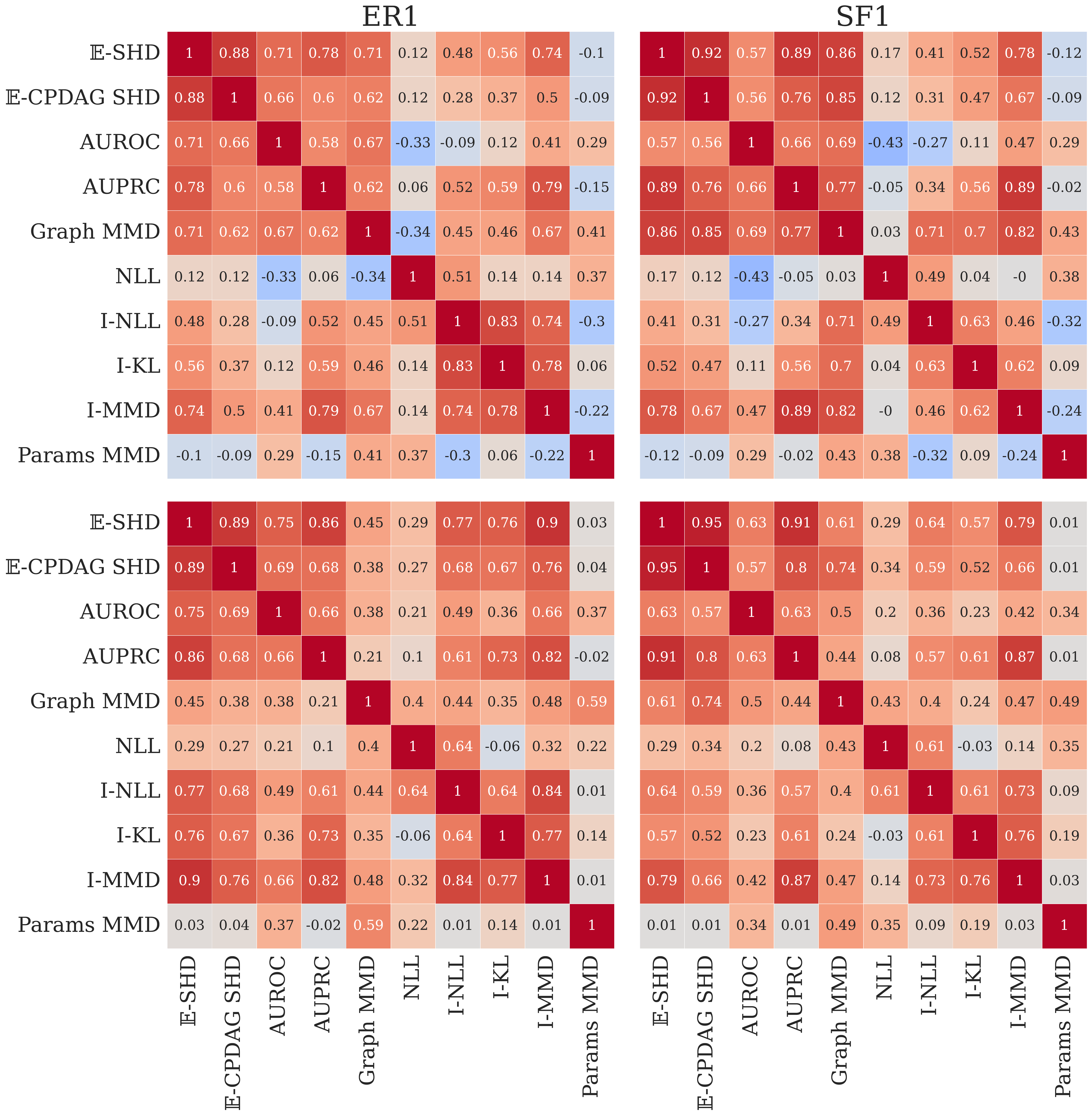}
	\caption{Spearman's rank correlation ceofficient between evaluation metrics with 1000 normalized training samples ($d=5$). The first and the second rows correspond to the non-identifiable and identifiable cases, respectively. Similar to the unnormalized case, All the graph-based metrics are correlated with each other and also the Graph MMD. Params MMD is also correlated with other metrics. Graph MMD and Params MMD are metrics that evaluate against the true posterior.}
	\label{fig:corr_normalized_1000}
\end{figure}

Recently, it has been shown that synthetic data might induce \emph{varsortability} bias, i.e. causal discovery algorithms take advantage of increasing marginal variance across the causal graph from root to leaf~\citep{reisach2021beware}. In order to account for this, we run all the methods wherein the marginal variance of each variable is roughly 1, and plot rank correlation (\cref{fig:corr_normalized_5,fig:corr_normalized_10,fig:corr_normalized_100,fig:corr_normalized_1000}). We observe that a similar pattern of correlation holds as before when the variables had different scales.


\subsection{Additional Results: Evaluation on Metrics}
\label{Appendix:metrics_results}
\begin{figure}[h]
	\centering
	\includegraphics[width=\textwidth]{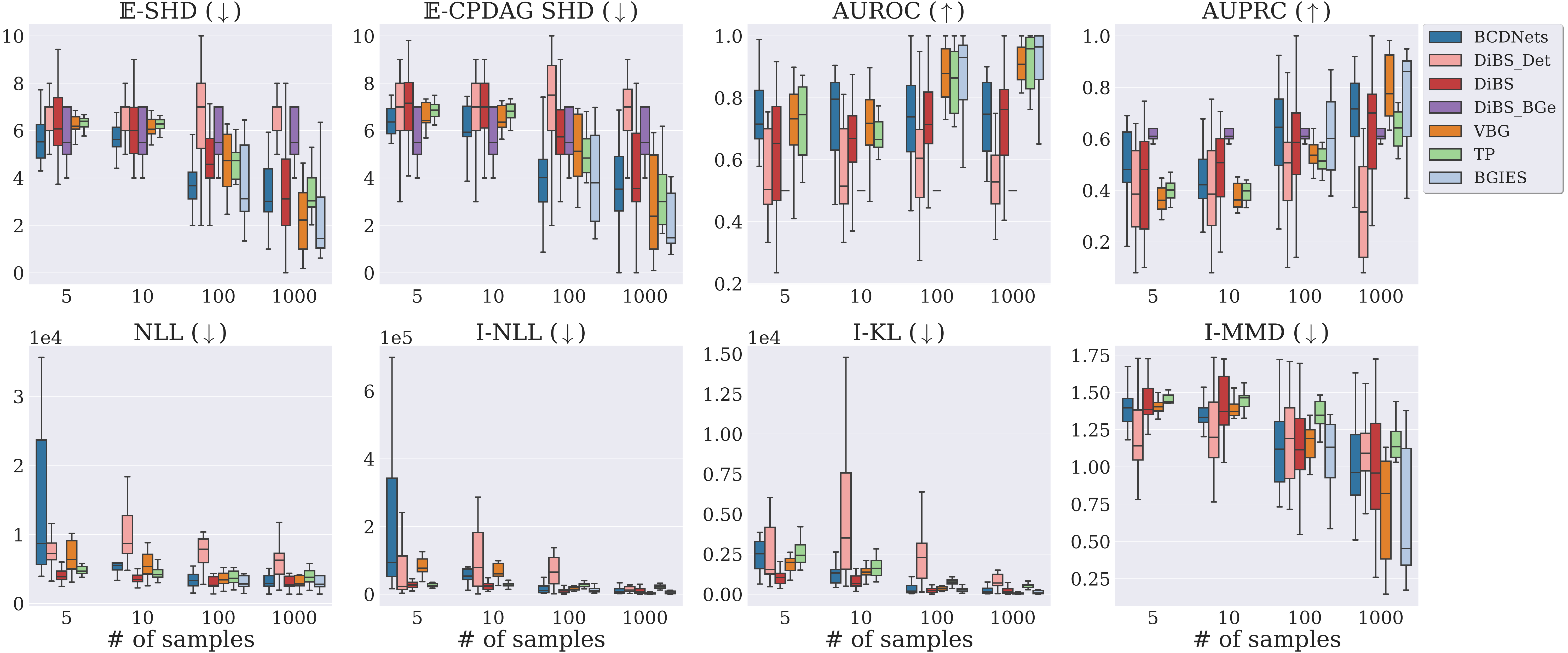}
	\caption{Evaluation of the models on ER2 graphs in the non-identifiable case ($d=5$). In low sample regimes, true posterior itself is evaluated to be worse on these metrics than their approximations.}
	\label{fig:er2-nonidentifiable}
\end{figure}

\begin{figure}[h]
	\centering
 	\includegraphics[width=\textwidth]{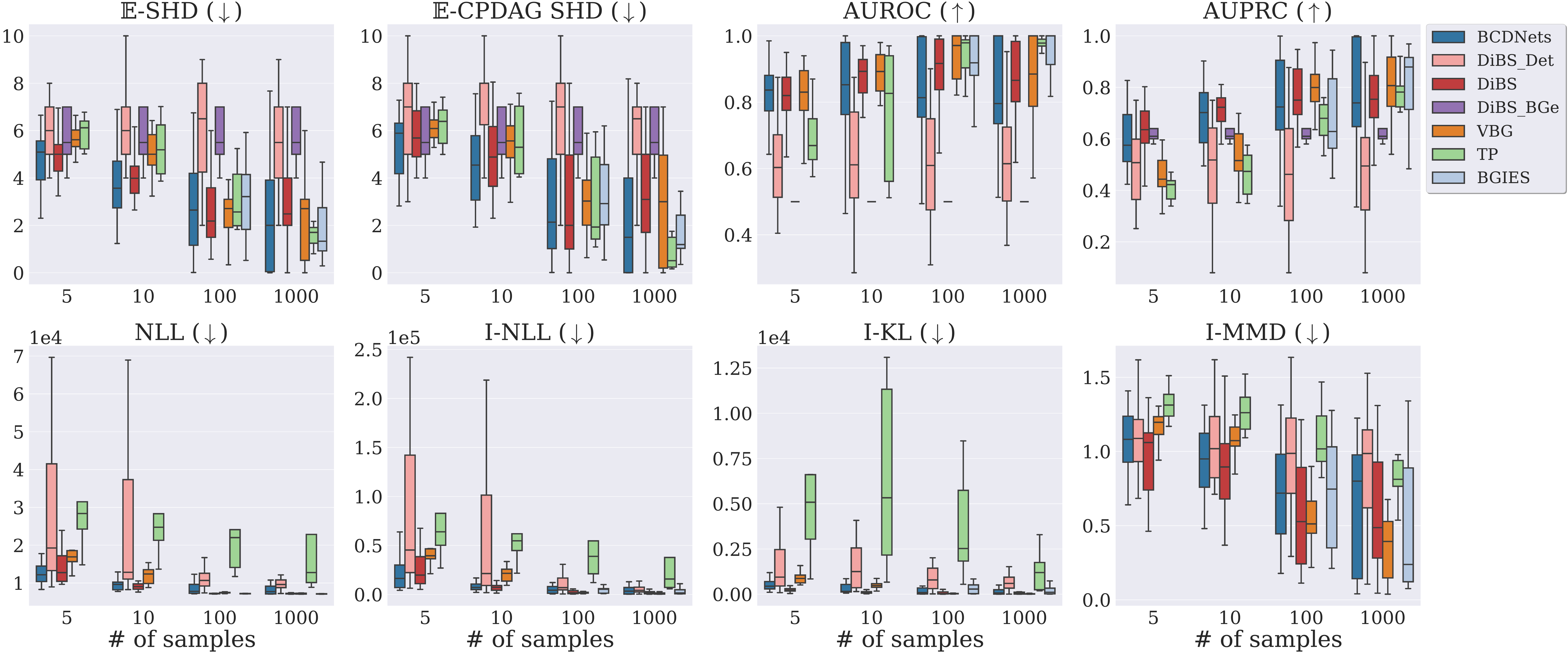}
 	\caption{Evaluation of the models on ER2 graphs in the identifiable case ($d=5$). In low sample regimes, true posterior itself is evaluated to be worse on these metrics than their approximations.}
 	\label{fig:er2-identifiable}
\end{figure}

\begin{figure}[h]
	\centering
 	\includegraphics[width=\textwidth]{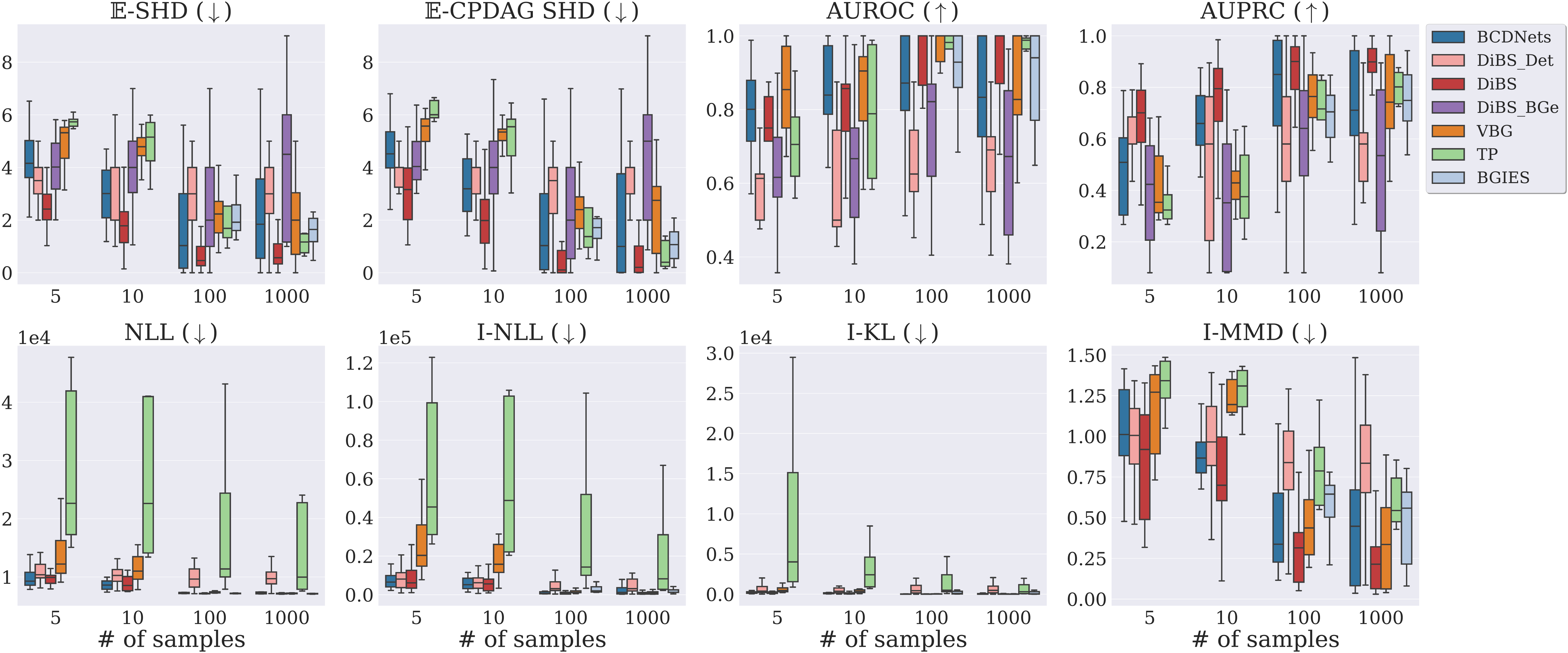}
 	\caption{Evaluation of the models on SF1 graphs in the identifiable case ($d=5$). In low sample regimes, true posterior itself is evaluated to be worse on these metrics than their approximations.}
 	\label{fig:sf1-identifiable}
\end{figure}

\begin{figure}[h]
	\centering
	\includegraphics[width=\textwidth]{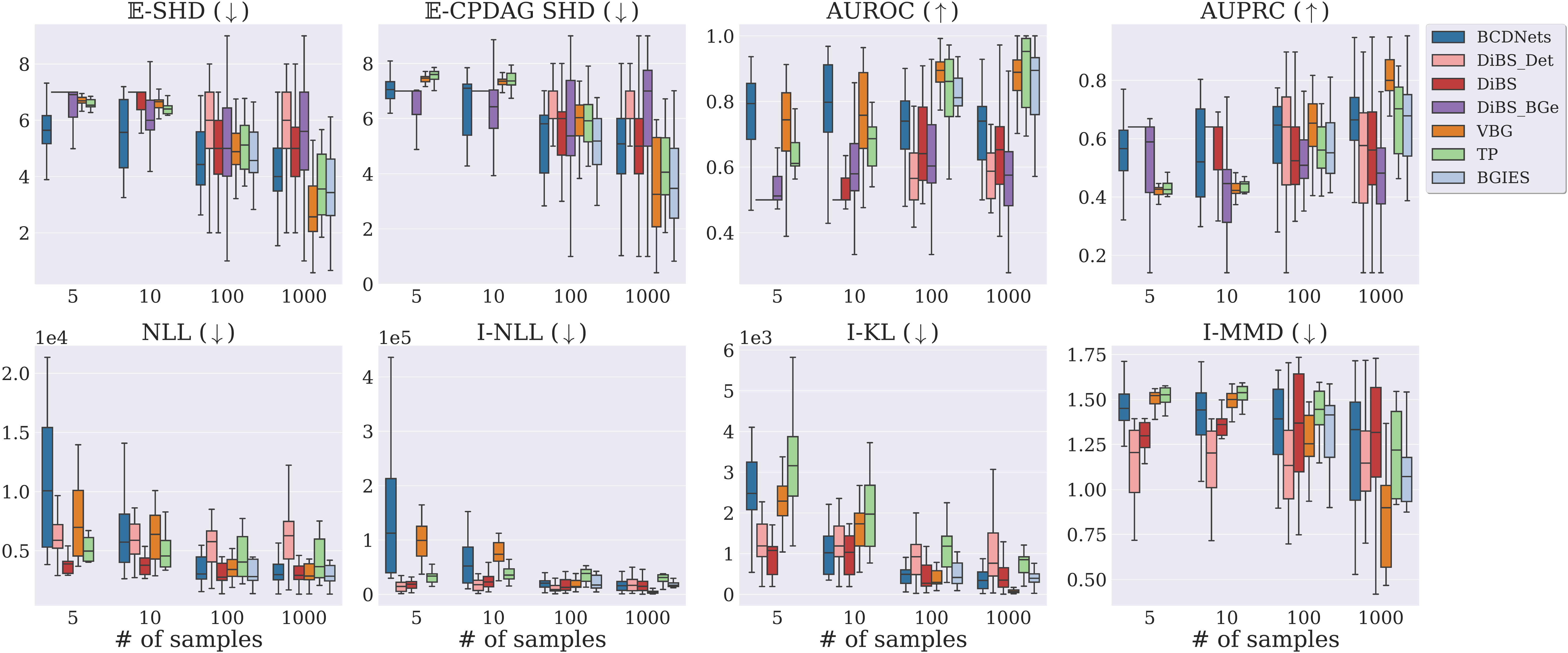}
	\caption{Evaluation of the models on SF2 graphs in the non-identifiable case ($d=5$). In low sample regimes, true posterior itself is evaluated to be worse on these metrics than their approximations.}
	\label{fig:sf2-nonidentifiable}
\end{figure}

\begin{figure}[h]
	\centering
 	\includegraphics[width=\textwidth]{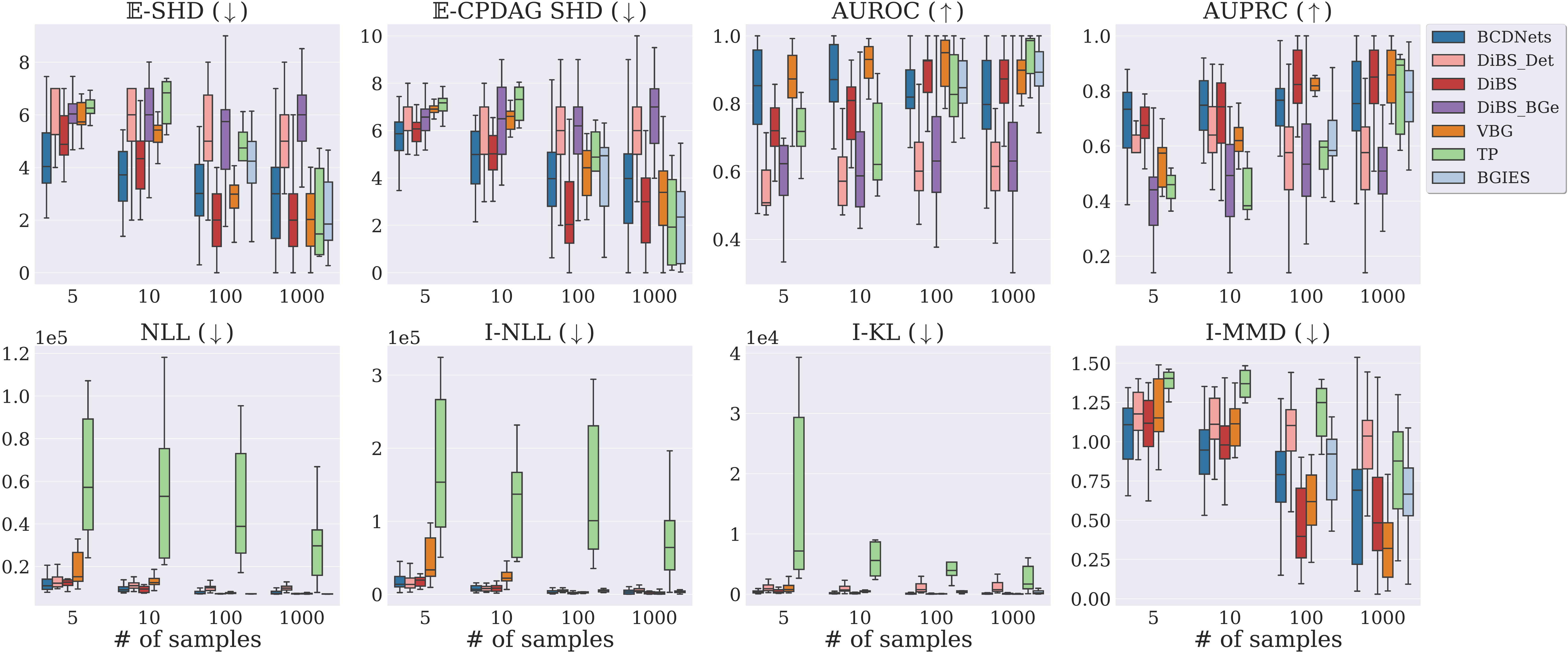}
 	\caption{Evaluation of the models on SF2 graphs in the identifiable case ($d=5$). In low sample regimes, true posterior itself is evaluated to be worse on these metrics than their approximations.}
 	\label{fig:sf2-identifiable}
\end{figure}

Here we report additional results. \cref{fig:er2-nonidentifiable,fig:er2-identifiable,fig:sf1-identifiable,fig:sf2-nonidentifiable,fig:sf2-identifiable} show the performance of models on different graph types in both non-identifiable and identifiable cases.



\subsection{Additional Results: Correlation Between Metrics}
\label{Appendix:corr_metrics}
\begin{figure}[h]
	\centering
	\includegraphics[width=\textwidth]{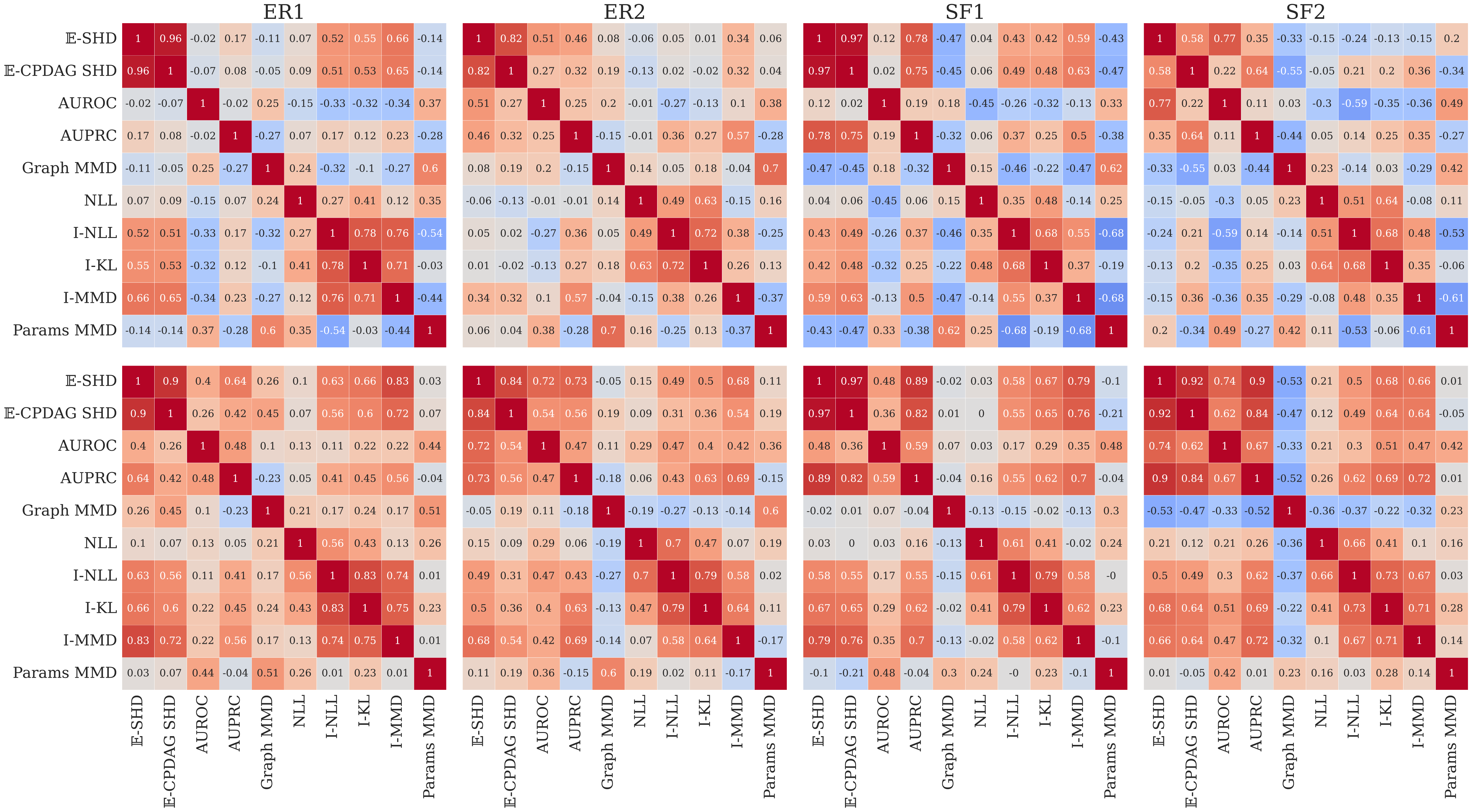}
	\caption{Spearman's rank correlation coefficient between evaluation metrics with 10 training samples ($d=5$). The first and the second rows correspond to the non-identifiable and identifiable cases, respectively. All the graph-based metrics are still not correlated with the Graph MMD, and Params MMD is still not correlated with any of the other metrics. Graph MMD and Params MMD are metrics that evaluate against the true posterior.}
	\label{fig:corr_10}
\end{figure}

\begin{figure}[h]
	\centering
	\includegraphics[width=\textwidth]{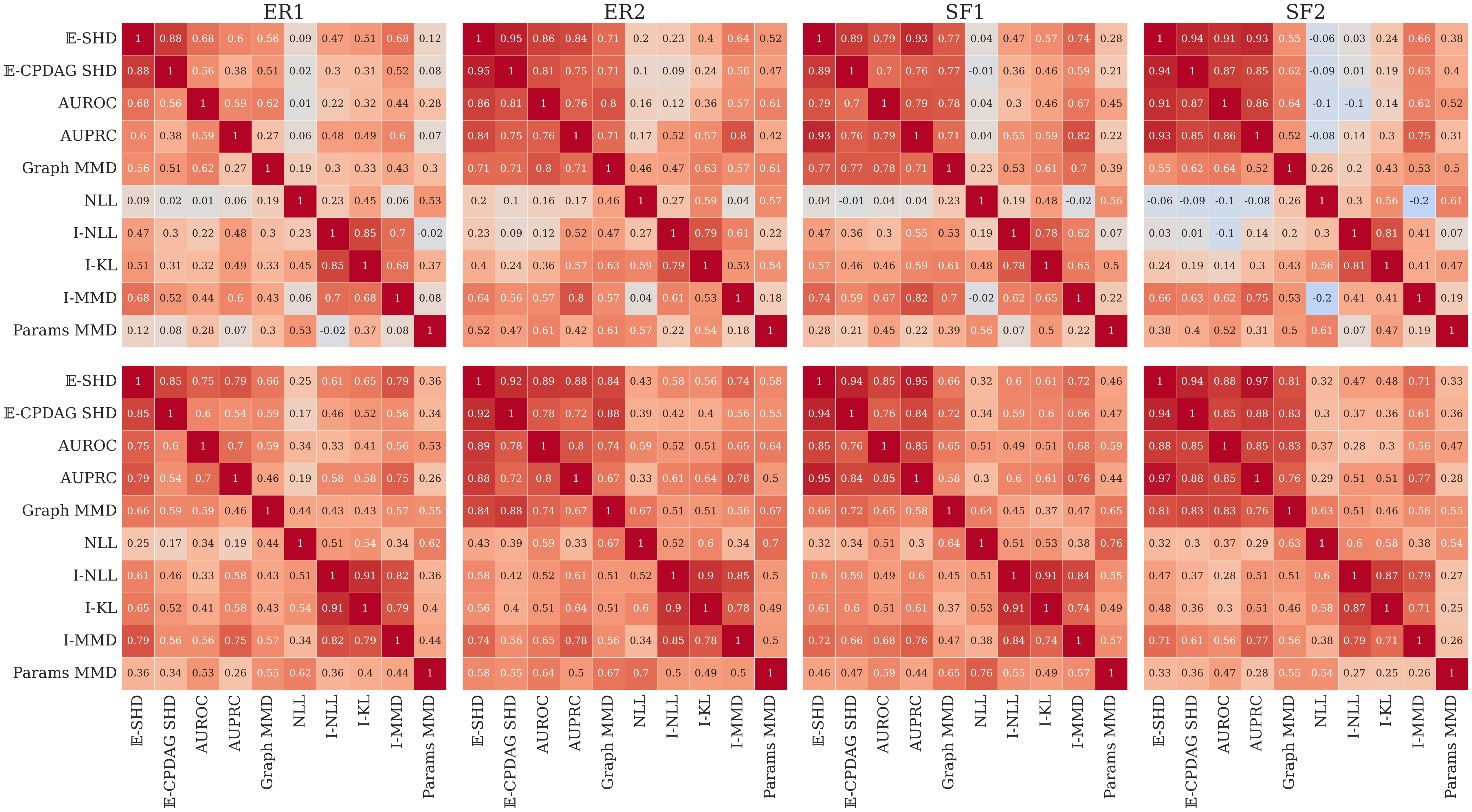}
	\caption{Spearman's rank correlation coefficient between evaluation metrics with 1000 training samples ($d=5$). The first and the second rows correspond to the non-identifiable and identifiable cases, respectively. All the graph-based metrics are starting to correlate with each other and also the Graph MMD. Params MMD is also start to correlate with other metrics. Graph MMD and Params MMD are metrics that evaluate against the true posterior.}
	\label{fig:corr_1000}
\end{figure}

\cref{fig:corr_10,fig:corr_1000} show the Spearman's rank correlation coefficient between evaluation metrics on 5-node graphs with 10 and 1000 samples, respectively. 

\subsection{Additional Results: Entropy and Comparison with True Posterior}
\label{appendix:entropy_models}
\begin{figure}[h]
	\centering
	\includegraphics[width=\textwidth]{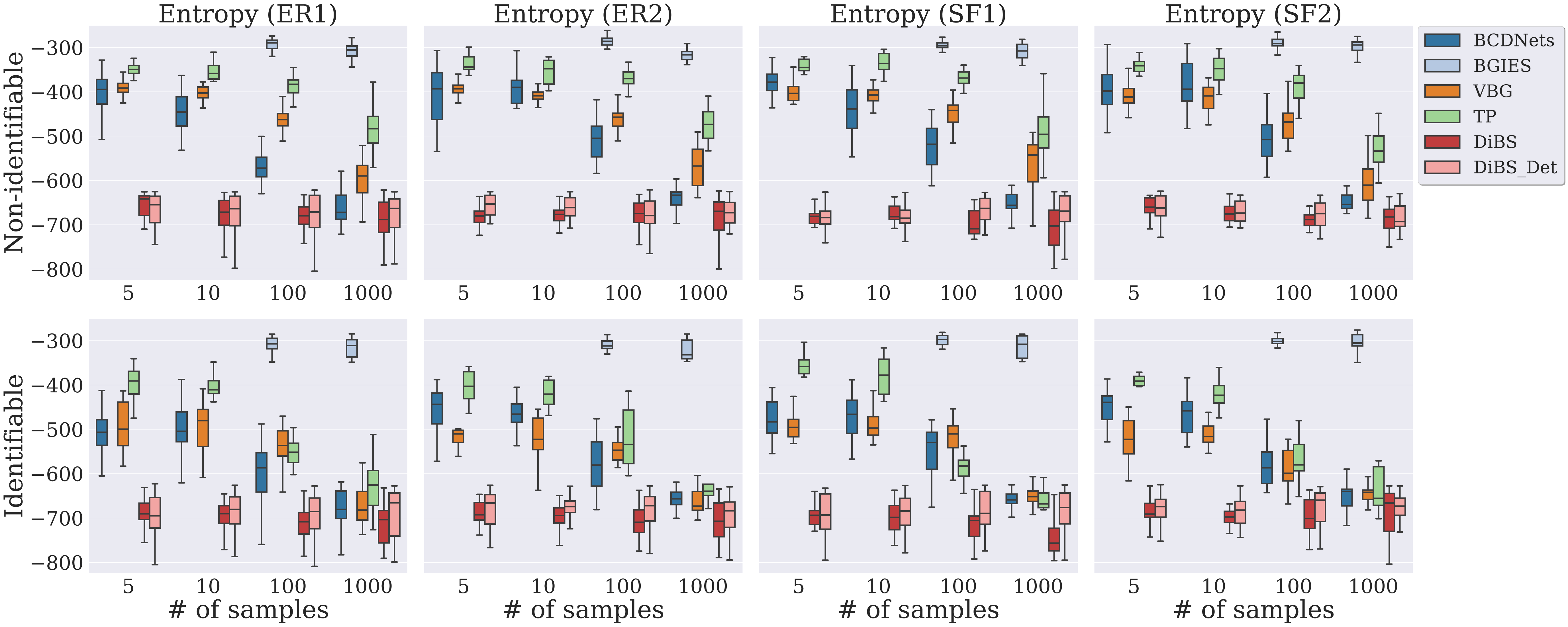}
	\caption{Entropy of the models on different graph types ($d=5$).}
	\label{fig:entropy}
\end{figure}

\begin{figure}[h]
	\centering
	\includegraphics[width=\textwidth]{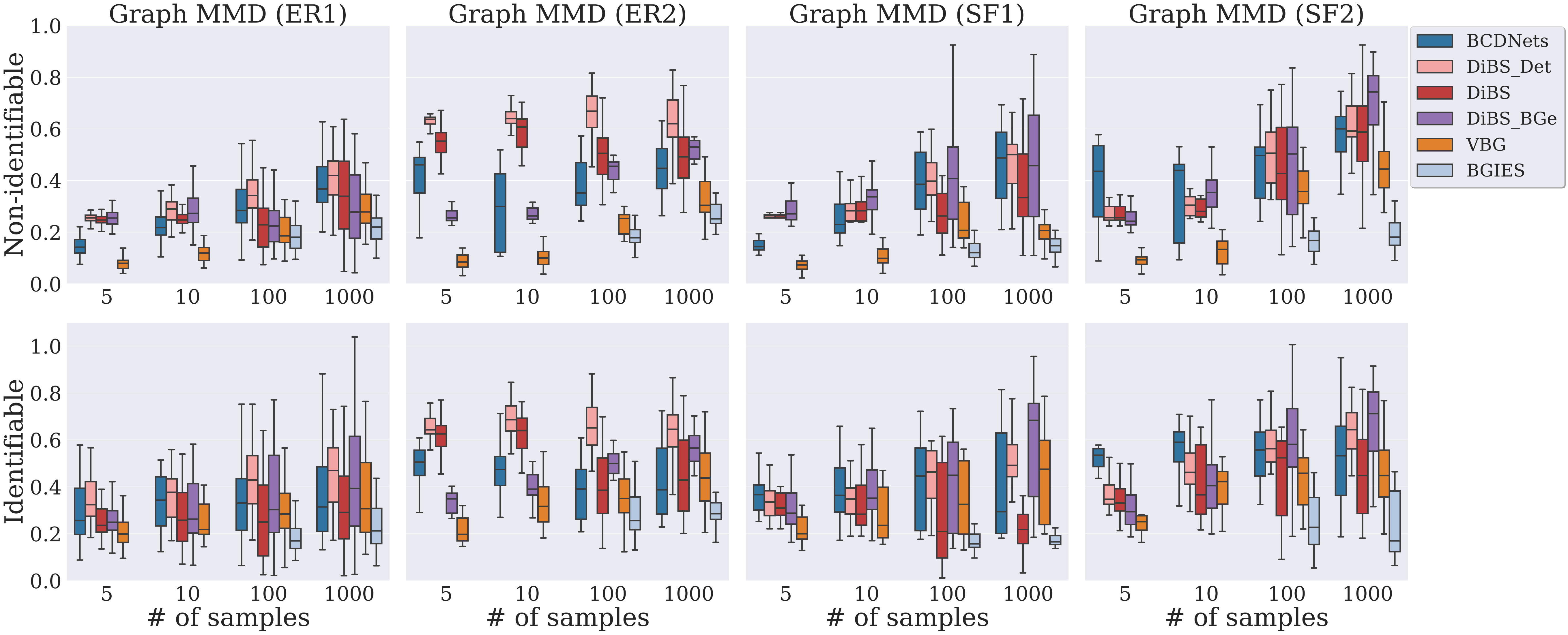}
	\caption{Graph MMD of the models on different graph types ($d=5$).}
	\label{fig:graph_mmd}
\end{figure}

\begin{figure}[h]
	\centering
	\includegraphics[width=\textwidth]{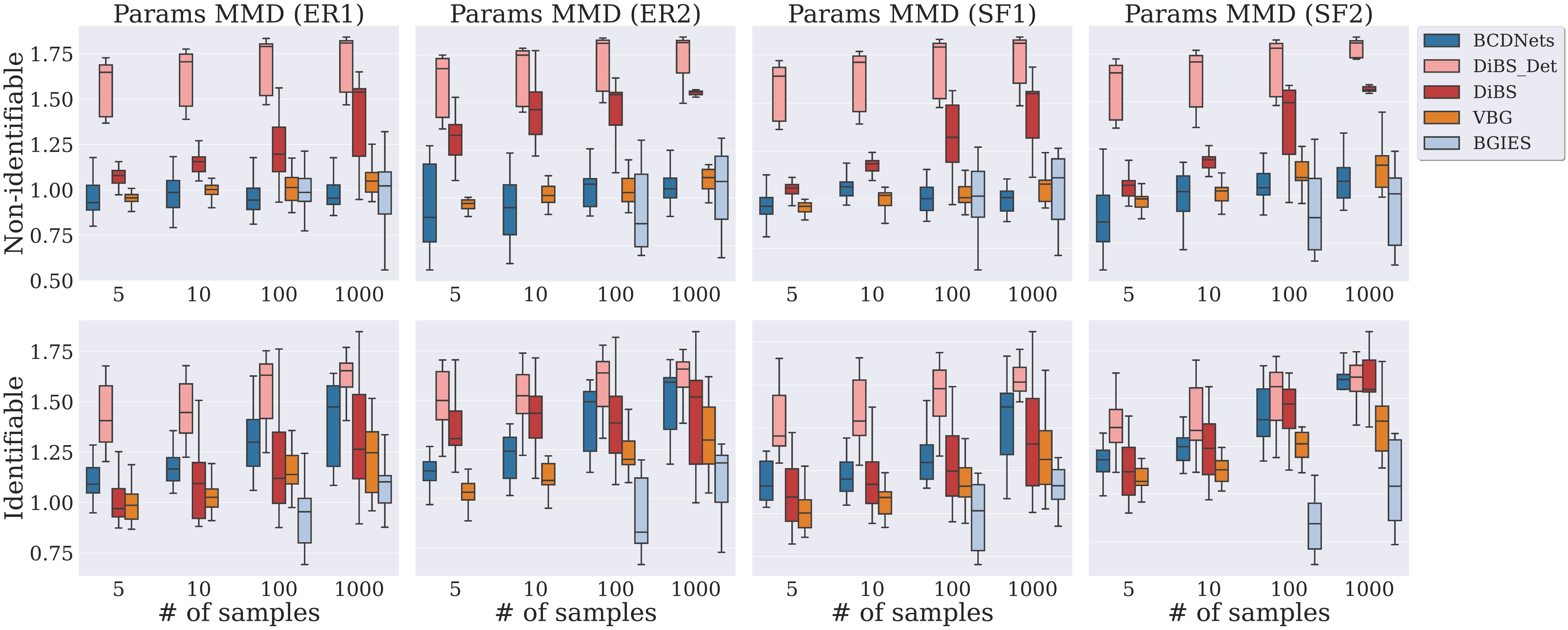}
	\caption{Params MMD of the models on different graph types ($d=5$).}
	\label{fig:params_mmd}
\end{figure}

\cref{fig:entropy} shows the entropy of the models on 5-node graphs with different graph types. \cref{fig:graph_mmd,fig:params_mmd} show the Graph MMD and Params MMD of the models on 5-node graphs with different types.

\end{document}